\documentclass[11pt]{article}
\usepackage[margin=1in]{geometry}
\usepackage[T1]{fontenc}
\usepackage{times}
\usepackage{amsmath,amssymb}
\usepackage{booktabs}
\usepackage{graphicx}
\usepackage{xcolor}
\usepackage[colorlinks=true,linkcolor=blue!50!black,citecolor=blue!50!black,urlcolor=blue!50!black]{hyperref}
\usepackage[numbers,sort&compress]{natbib}
\usepackage{microtype}
\usepackage{enumitem}
\usepackage[section]{placeins}
\usepackage{tikz}
\usetikzlibrary{arrows.meta,positioning,fit,calc}
\setlist{nosep,leftmargin=*}

\newcommand{\NCfg}{928}
\newcommand{\NInt}{1{,}405{,}500}
\newcommand{\NCfgVone}{72}
\newcommand{\NCfgHtw}{816}
\newcommand{\NIntHtw}{1{,}224{,}000}
\newcommand{\NCfgAsc}{39}
\newcommand{\NIntAsc}{79{,}200}
\newcommand{\IOIkFourSucc}{12.3}
\newcommand{\IOIkFourMax}{31.3}
\newcommand{\IOIkFourN}{210}
\newcommand{\IOIkFourHPC}{0}

\newcommand{\hpc}{\textsc{HPC}}
\newcommand{\LD}{\mathrm{LD}}

\title{Right Answer, Wrong Mechanism:\\Detecting Pernicious Divergence in Causal Interventions}
\author{
Beiming Liu\thanks{Equal contribution (co-first authors).}\\
China Financial Certification Authority (CFCA), China\\
\texttt{lbm21@tsinghua.org.cn}
\and
Minjie Chen\footnotemark[1]\\
PetroChina Southwest Oil and Gasfield Company, China\\
\texttt{auturnncrow@gmail.com}
}
\date{}

\begin{document}
\maketitle

\begin{abstract}
Causal interventions such as activation patching and distributed alignment search (DAS) are the main tool for making mechanistic claims about neural networks. Recent work showed that these interventions routinely push representations off the model's natural distribution, and that such divergence is sometimes harmless and sometimes \emph{pernicious}: it can recruit pathways the model never uses on natural inputs, so that an intervention produces the expected answer through the wrong mechanism. No method currently tells the two cases apart. We make this question testable by planting hidden pathways inside pretrained language models. The pathways are silent on every benchmark prompt by construction, so model behaviour is unchanged, and which interventions depend on them is known exactly. Across \NCfg{} configurations and \NInt{} interventions on four models from 124M to 7B parameters (GPT-2 small, Pythia-1.4B, Pythia-6.9B and Qwen2.5-7B, three seeds for the larger models), we find four things.
(i) Nearest-neighbour and local-PCA distances at the intervention site, as used in prior work, are \emph{below chance} at picking out interventions that give the right answer through a planted pathway on the gender task for every model (AUROC 0.21--0.46), and never exceed 0.62 on any task.
(ii) Hidden-Pathway Contribution (\hpc{}) is a label-free downstream test. It clamps downstream units to the regime seen in natural runs with the same output and measures how much of the decision disappears. Among successful interventions it identifies wrong-mechanism successes with AUROC 0.82--1.00 for every model when the pathway shows up as unit-level out-of-regime activity. When every unit of the pathway stays within its natural range, range clamping drops to 0.39--0.70 on the three larger models; first projecting downstream activity onto a local natural manifold recovers 0.73--0.85 on both Pythia models and on Qwen2.5-7B SVA, but not on Qwen2.5-7B gender (0.48) or GPT-2 gender (0.57). The in-range case remains open.
(iii) Optimised interventions actively seek hidden pathways, at every scale. On the gender task, 81--100\% of the successes of DAS trained with unit-visible pathways present are pathway-dominated, and DAS restricted away from the genuine causal subspace still succeeds, mostly through planted pathways. On GPT-2, a downstream on-manifold penalty cuts the pathway-dominated share of DAS successes from nearly all to under 5\%.
(iv) In unmodified GPT-2 and Pythia models, successful interventions show almost no unit-level out-of-regime reliance ($\leq$0.8\%), and none of the 2{,}232 successes of restricted DAS in Pythia-6.9B with a single component excluded does. In unmodified Qwen2.5-7B, restricted DAS at the final token produces the first candidates: about 30\% of its SVA successes depend on out-of-regime downstream activity. We bound what these checks can see with a positive control.
\end{abstract}

\section{Introduction}

Much of mechanistic interpretability rests on one experimental primitive: change an internal representation and observe the output. Activation patching~\citep{vig2020causal,meng2022rome,wang2023ioi}, interchange interventions and distributed alignment search (DAS)~\citep{geiger2021causal,geiger2024das,wu2023boundless}, and mean-difference patching~\citep{feng2024binding} all follow this template. Their results are read as evidence about the model's \emph{natural} mechanism, which assumes the intervened state is one the model could plausibly be in.

\citet{grant2026divergent} showed that this assumption routinely fails. Common interventions move representations off the natural distribution. That divergence can be \emph{harmless}, lying in directions the downstream computation ignores, or \emph{pernicious}, activating ``hidden pathways'' that produce the intended behaviour through a mechanism the model never uses. Earlier, \citet{makelov2024subspace} constructed a concrete instance of the pernicious case, and \citet{wu2024reply} argued about how often such constructions arise in practice. \citet{grant2026divergent} leave open ``a principled method for classifying harmful divergence''. Without one, a researcher cannot tell whether a successful intervention supports a claim about the natural mechanism.

Two obstacles make the question hard. First, there is no ground truth: in a real network we do not know which interventions succeed for the right reason. Second, the obvious candidate signal is how far the intervened representation lies from the data manifold at the intervention site. By construction, that signal responds to harmless and pernicious divergence alike. This paper addresses both obstacles.

\paragraph{Contributions.}
\begin{itemize}
\item \textbf{A planted-pathway benchmark in pretrained language models} (\S\ref{sec:bench}). We add hidden pathways to GPT-2 small, Pythia-1.4B, Pythia-6.9B and Qwen2.5-7B. They are silent on all benchmark prompts, so behaviour is unchanged, but they fire when an intervention pushes the downstream state out of its natural regime. Four families range from new silent units to drive spread over 128 existing neurons, each clipped to its natural range. Switching the pathways off gives an exact label for every intervention.
\item \textbf{Site-level distances do not identify wrong-mechanism successes} (\S\ref{sec:res-main}). Among interventions that produce the counterfactual answer, nearest-neighbour and local-PCA distances are below chance on the gender task for every model, and at best weakly informative elsewhere: interventions that succeed through a planted pathway are, if anything, \emph{closer} to the natural site manifold. Mahalanobis distance and the Algorithm~1 of \citet{grant2026divergent} are inconsistent across models, tasks and pathway families.
\item \textbf{Hidden-Pathway Contribution} (\S\ref{sec:hpc}). This label-free detector asks a causal question downstream: how much of the decision survives when downstream units are clamped to the regime of natural runs with the same output? It works at every scale when pathways are visible as unit-level out-of-regime activity. A variant that first projects onto the local natural manifold partly closes the in-range gap on the Pythia models and on Qwen2.5-7B SVA, but no variant is reliable on in-range pathways, which we present as the open hard case.
\item \textbf{Optimised interventions find hidden pathways} (\S\ref{sec:res-illusion}). On the gender task, for the three unit-visible families and every model, DAS trained on a model with planted pathways routes most of its successes through them. Restricted to exclude the genuine causal subspace, it still reports success, mostly through planted pathways. On GPT-2, a downstream on-manifold penalty redirects unrestricted DAS to the natural mechanism, but does not remove the false successes of restricted DAS.
\item \textbf{A bounded prevalence check} (\S\ref{sec:res-natural}). In unmodified GPT-2 and Pythia models, successful interventions almost never show unit-level out-of-regime reliance, even when restricted DAS succeeds often. In unmodified Qwen2.5-7B, restricted DAS at the final token yields candidate wrong-mechanism successes. A positive control shows what this check can and cannot detect.
\end{itemize}

\begin{figure}[t]
\centering
\begin{tikzpicture}[
  box/.style={draw, rounded corners=2pt, minimum width=1.25cm, minimum height=0.62cm, font=\footnotesize, align=center},
  trap/.style={box, draw=red!70!black, fill=red!6, dashed},
  arr/.style={-{Stealth[length=2mm]}, thick},
  lab/.style={font=\scriptsize, align=center}]
\node[box, fill=gray!10] (x) {prompt};
\node[box, right=0.55cm of x, fill=blue!8] (site) {layer $L$\\site $h$};
\node[box, right=0.8cm of site] (mlp) {MLPs\\$L{+}1..N$};
\node[trap, below=0.55cm of mlp] (tr) {hidden\\pathway};
\node[box, right=0.8cm of mlp, fill=gray!10] (y) {logit diff.};
\draw[arr] (x) -- (site);
\draw[arr] (site) -- node[lab, above] {$\hat h$} (mlp);
\draw[arr] (mlp) -- (y);
\draw[arr, red!70!black, dashed] (site) |- (tr);
\draw[arr, red!70!black, dashed] (tr) -| (y);
\node[lab, below=0.12cm of tr, red!70!black] {planted; silent on all benchmark prompts};
\node[lab, text width=2.6cm, above=0.12cm of site, gray!40!black] (sl) {site-level measures\\look only at $\hat h$};
\node[box, draw=blue!60!black, fill=blue!4, above=0.75cm of mlp, xshift=1.3cm, text width=4.3cm] (h) {\textbf{HPC}: clamp downstream units to the natural regime of runs with the same output; measure the lost margin};
\draw[arr, blue!60!black] (h.south -| mlp.north) -- (mlp);
\end{tikzpicture}
\caption{\textbf{Setting.} An intervention replaces the site representation $h$ by $\hat h$. It may reach the right answer through the natural computation or through a pathway that natural inputs never activate. Site-level divergence looks only at $\hat h$. \hpc{} tests downstream whether the decision depends on activity outside the natural regime. In our benchmark the hidden pathway is planted, so the answer is known.}
\label{fig:setting}
\end{figure}

\section{Related work}

\paragraph{Causal interventions and their pitfalls.} \citet{zhang2024patching} and \citet{heimersheim2024patching} review activation patching and discuss metric and corruption choices. Circuit-discovery methods build on it~\citep{conmy2023acdc}. Causal scrubbing~\citep{chan2022scrubbing} resamples activations to keep tests on-distribution, and optimal ablations~\citep{li2024optimal} choose ablation values that minimise distortion. \citet{makelov2024subspace} showed that subspace patching can succeed through a dormant parallel pathway; \citet{wu2024reply} disputed how common this is in practice. \citet{grant2026divergent} showed that divergence is common across patching, DAS and SAE reconstructions and analysed when it is harmless. They proposed a counterfactual-latent loss, which shrinks divergence overall but does not target its pernicious part. \citet{sutter2025nonlinear} showed that sufficiently expressive alignment maps can align any network to any algorithm. \citet{vaidyanathan2026curse} showed that patching estimates contain interaction effects that grow with the distance between clean and patched activations. Downstream self-repair~\citep{mcgrath2023hydra} is a related complication for any method that reasons about downstream effects. Our work supplies the missing evaluation: ground truth for which interventions are pernicious, and detectors scored against it.

\paragraph{Keeping interventions on-manifold.} \citet{luo2026glp} train a diffusion model over activations and project steered activations back onto the learned manifold. Like Mahalanobis or nearest-neighbour scores~\citep{lee2018mahalanobis,sun2022knn}, this assesses the representation \emph{at the site}. A related out-of-distribution problem affects perturbation-based attribution~\citep{hooker2019roar,hase2021ood}.

\paragraph{Ground truth for interpretability.} Compiled transformers~\citep{lindner2023tracr}, semi-synthetic models with known circuits~\citep{gupta2024interpbench} and standardised benchmarks~\citep{mueller2025mib} evaluate circuit and alignment methods. Instead of building a model with a known circuit, we make a minimal edit to a pretrained model. The added pathways have no effect on the benchmark prompts, so the natural computation remains that of the real model.

\section{Problem setup}
\label{sec:setup}

\paragraph{Decisions and interventions.} Consider a binary decision measured by the logit difference $\LD(x)=\ell_{a_1}(x)-\ell_{a_0}(x)$ between two answer tokens. A causal variable takes two values, corresponding to classes $c\in\{0,1\}$, and target and source prompts $(x_t,x_s)$ differ only in that variable. An intervention at layer $L$ and position $p$ replaces the residual stream $h=h^p_L(x_t)$ by $\hat h$. It is \emph{successful} if $\operatorname{sign}\LD(x_t;\hat h)$ matches the class of $x_s$.

\paragraph{Pernicious divergence.} We call an intervention \emph{pernicious for natural-mechanism claims} if its decision depends on downstream computation that does not occur in natural runs producing the same output. This follows \citet{grant2026divergent}, for whom harm is relative to the claim. Ours is the most common claim: that the intervened representation is used by the model's own mechanism. In real models this property cannot be observed, which motivates the benchmark below.

\section{A planted-pathway benchmark}
\label{sec:bench}

\paragraph{Tasks.} We use two tasks with paired prompts that differ in one variable and share fixed answer tokens:
\begin{itemize}
\item \textbf{SVA}, subject--verb agreement~\citep{linzen2016agreement,finlayson2021agreement}: ``The keys near the cabinet'' $\to$ \emph{are}/\emph{is}. 47 noun pairs, 8 prepositions, singular and plural attractors.
\item \textbf{Gender} pronouns~\citep{vig2020causal}: ``The king from Paris said that'' $\to$ \emph{he}/\emph{she}. 20 noun pairs, 20 locations, 12 verbs.
\end{itemize}
A third task, indirect object identification (\textbf{IOI})~\citep{wang2023ioi}, has prompt-dependent answers, so we use it only in the unmodified-model analysis. We keep prompt pairs the model answers correctly with $|\LD|>1$.

\paragraph{Models and sites.} We use GPT-2 small~\citep{radford2019gpt2}, Pythia-1.4B and Pythia-6.9B~\citep{biderman2023pythia}, and Qwen2.5-7B~\citep{qwen2025qwen25}, all run with TransformerLens~\citep{nanda2022transformerlens} at fixed model revisions. Sites are the subject token at three depths and the final token at one depth:
GPT-2 (12 layers), subject at $L\in\{2,4\}$ and final token at $L{=}8$;
Pythia-1.4B (24 layers), subject at $\{4,8,12\}$ and final at $16$;
Pythia-6.9B (32 layers), subject at $\{6,12,18\}$ and final at $24$;
Qwen2.5-7B (28 layers), subject at $\{6,12,18\}$ and final at $20$.
GPT-2 configurations use seed 0; the three larger models use seeds 0, 1 and 2. Appendix~\ref{app:impl} lists all budgets.

\paragraph{Planted pathways.} Fix a layer $T>L$. Let $\tilde x_T(x)$ be the normalised input of the MLP at layer $T$ at the final position, and let $u$ be a unit direction. A pathway unit computes
\begin{equation}
a(x)=\operatorname{ReLU}\!\big(u^\top\tilde x_T(x)-b\big),\qquad b=\max_{x\in\mathcal D}u^\top\tilde x_T(x)+\delta,\quad \delta=\eta\,\operatorname{std}_{x\in\mathcal D}(u^\top\tilde x_T(x)),\ \eta\sim U(0.05,0.8),
\end{equation}
where $\mathcal D$ is the set of benchmark prompts. So $a\equiv0$ on every benchmark prompt, and we assert that the model's decisions on them are unchanged.

Each configuration plants 24 units:
\begin{itemize}
\item Half of the directions $u$ lie in the top-64 principal subspace of $\tilde x_T$, reachable by overshooting natural directions. The other half lie in its bottom-200 subspace, reachable only off-manifold, in the style of \citet{makelov2024subspace}.
\item Half of the units are \emph{confirmatory}. For a top-subspace unit this means pushing the answer that its input direction correlates with. For a bottom-subspace unit, whose direction carries no class information, the sign is random. The remaining units push in a random direction.
\end{itemize}
Each unit's output is calibrated so that an overshoot of one natural standard deviation moves $\LD$ by about 3 (by about 30 before clipping for in-range; see below). The four families differ in where the output goes:
\begin{itemize}
\item \textbf{silent}: new units that write along the unembedding difference of the answers (confirmatory) or along a random token's unembedding;
\item \textbf{hijack}: extra drive into one existing MLP neuron at layer $T$ with a large direct effect on $\LD$;
\item \textbf{distributed}: the same drive spread over 64 such neurons;
\item \textbf{in-range}: drive spread over 128 neurons, each clipped so that it never leaves its natural range, taken as the 0.1--99.9\% quantiles over all benchmark prompts. Only the \emph{combination} of neuron values is unnatural. The gain is ten times larger to offset the clipping.
\end{itemize}
The first three families make the pathway visible as unit-level out-of-regime activity, which is exactly what \hpc{} tests for. They are therefore a sanity check and an upper bound. The in-range family breaks this assumption and is the informative test. Pathways sit at $T\in\{L{+}2,L{+}4\}$ (for GPT-2, $T<12$).

\paragraph{Ground truth.} For every intervention we compare $\LD$ with the pathways on ($\LD$) and off ($\LD_{\mathrm{off}}$). We use two labels:
\begin{itemize}
\item \emph{Pathway-dominated}: the pathways supply more than half of the decision margin, i.e.\ $\operatorname{sign}(\LD)\,(\LD-\LD_{\mathrm{off}}) > |\LD|/2$. We use this label over all interventions, where many are unsuccessful.
\item \emph{Wrong-mechanism success}: the intervention is successful and its decision flips when the pathways are off. This is the label that matters for interpretation, used among successful interventions.
\end{itemize}

\paragraph{Interventions.} Each configuration evaluates 1{,}500 interventions on held-out pairs, drawn in equal numbers from the following types:
\begin{itemize}
\item \textbf{A}: full activation patch, $\hat h = h_s$;
\item \textbf{C}: mean-difference patching $\hat h=h_t+\alpha(\mu_{s}-\mu_{t})$, with $\alpha\sim U(0.1,5)$, plus an equally sized group with $\alpha\sim U(0,3)$ (labelled ``steering'' in the code);
\item \textbf{D}: 1-D DAS~\citep{geiger2024das} trained on the unmodified model, with the interchange scaled by $U(0.2,4)$;
\item \textbf{E}: DAS mixed with a random natural direction at up to $85^\circ$;
\item \textbf{G}: DAS trained on the model \emph{with} pathways;
\item \textbf{H}: \emph{restricted} DAS, also trained with pathways, constrained to the orthogonal complement of the top-$k$ principal components of the site ($k{=}32$ unless stated) and of the direction from D.
\end{itemize}
H is an illusion probe: if the genuine signal lies in the excluded subspace, any success must come from elsewhere.

\section{Detectors}
\label{sec:hpc}

\paragraph{Site-level baselines.} These score $\hat h$ against the natural site representations:
\begin{itemize}
\item \textbf{kNN}: distance to the nearest natural neighbour~\citep{sun2022knn}.
\item \textbf{Mahalanobis}~\citep{lee2018mahalanobis}: computed in a 200-dimensional PCA basis, plus the out-of-subspace residual scaled by the smallest retained variance.
\item \textbf{LPCA}: the local-PCA residual of \citet{grant2026divergent}, i.e.\ the distance from the tangent space of the 30 nearest neighbours, keeping 95\% of variance.
\item \textbf{Algorithm~1} of \citet{grant2026divergent}, reimplemented. It extracts the off-manifold component $v$ of $\hat h$ relative to the local PCA of the predicted class, adds $v$ to 32 natural evaluation prompts, and reports the largest change in $\LD$.
\end{itemize}

\paragraph{Downstream off-manifold (Down-OM).} A non-causal downstream baseline. For each analysed MLP layer $l>L$, compute the relative residual of the final-position activation from the class-conditional natural PCA subspace (99\% of variance, at most 256 components) of the predicted class, and average over layers.

\paragraph{Hidden-Pathway Contribution.} For each downstream MLP neuron $j$ (final position, layers $>L$) and class $c$, record over natural prompts of class $c$ the regime $[q^c_{0.001}(j),q^c_{0.999}(j)]$ and the frequency with which $|a_j|>0.05$. Given an intervention with predicted class $c^*$, rerun the forward pass with every downstream neuron clamped to the regime of $c^*$. The clamps are applied layer by layer, so later layers see clamped inputs. Let $\Pi$ denote this clamping and $\LD_\Pi$ the resulting logit difference. Then
\begin{equation}
\hpc(\hat h)=\frac{\max\!\big(0,\ \operatorname{sign}(\LD)\,(\LD-\LD_\Pi)\big)}{\operatorname{median}_{x\in\mathcal D}|\LD(x)|},
\end{equation}
the fraction of a typical natural margin lost once downstream activity is restricted to what natural runs with the same output exhibit. \hpc{} needs only natural prompts and the model, no labels. We use four variants:
\begin{itemize}
\item \textbf{HPC-S} (strict) clamps only neurons active in fewer than 1\% of natural class-$c^*$ runs.
\item \textbf{HPC-L} (loose) clamps every neuron to its regime.
\item \textbf{HPC-PCA} first projects each layer's activation onto the class-$c^*$ natural PCA subspace used by Down-OM, then clamps.
\item \textbf{HPC-Loc} first projects each layer's activation onto the local PCA of its 30 nearest natural class-$c^*$ activations, then clamps.
\end{itemize}
Harmless null-space divergence leaves downstream activity unchanged, so \hpc{} is zero for it by construction. Planted units of the silent family are treated like any other downstream unit. Their natural regime is $\{0\}$, so HPC-S and HPC-L switch them off, which is close to the operation that defines the label. This is why the silent family serves only as a sanity check. The two projection variants can in principle see combination-level anomalies that no single unit's range reveals, which is what the in-range family requires.

\section{Results}
\label{sec:results}

\subsection{A synthetic sanity check}
We first extend the synthetic setting of \citet{grant2026divergent}. A two-layer MLP classifies 10 classes defined by two correlated latent variables, and we plant silent units and a Makelov-style dormant pathway. Table~\ref{tab:synthetic} shows the pattern that recurs below. Site distances flag harmless null-space divergence as readily as pernicious divergence, with 93--100\% false positives at 90\% recall. Algorithm~1 is much better, and \hpc{} is best.

\begin{table}[!htbp]
\centering
\small
\caption{\textbf{Synthetic setting} (5 seeds, 3{,}000 interventions each). AUROC for detecting trap-dependent interventions, over all and over successful interventions, and false-positive rate (\%) on harmless null-space divergence at the threshold that catches 90\% of pernicious cases. $^\dagger$\citet{grant2026divergent}.}
\label{tab:synthetic}
\begin{tabular}{l ccc}
\toprule
Detector & AUROC (all) & AUROC (successful) & FPR null-space \\
\midrule
site kNN & 0.651$\pm$0.034 & 0.586$\pm$0.034 & 94.8 \\
site Mahalanobis & 0.803$\pm$0.038 & 0.751$\pm$0.041 & 99.6 \\
site local-PCA$^\dagger$ & 0.681$\pm$0.030 & 0.618$\pm$0.031 & 93.2 \\
Algorithm~1$^\dagger$ & 0.837$\pm$0.016 & 0.868$\pm$0.012 & 4.0 \\
HPC (ours) & 0.931$\pm$0.007 & 0.937$\pm$0.005 & 0.0 \\
\bottomrule
\end{tabular}
\end{table}

\subsection{Site-level distance does not identify wrong-mechanism successes}
\label{sec:res-main}

\begin{figure}[!htbp]
\centering
\includegraphics[width=\linewidth]{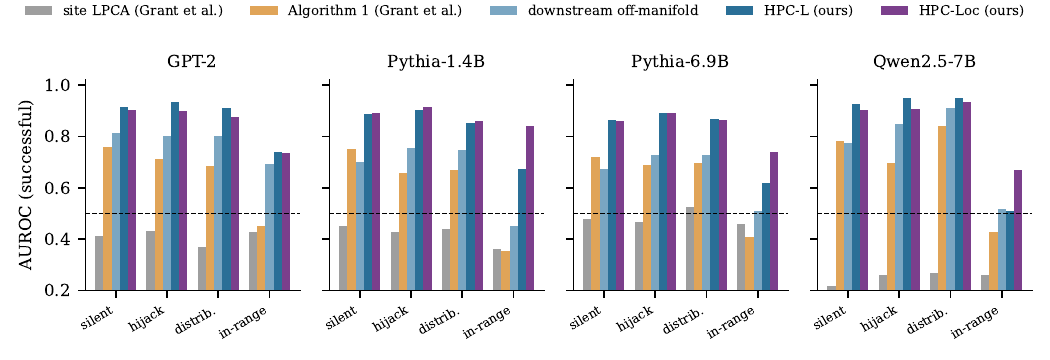}
\caption{AUROC for wrong-mechanism successes, averaged over the two tasks, by model and pathway family. Site-level LPCA is near or below chance throughout. HPC-L is near the top for unit-visible families on every model. On in-range pathways, HPC-Loc is better than HPC-L on the three larger models; the averages hide chance-level performance of all detectors on Qwen2.5-7B gender (Table~\ref{tab:auroc_success}).}
\label{fig:bars}
\end{figure}

Table~\ref{tab:auroc_success} scores the detectors on the question that matters for interpretation: among interventions that \emph{succeeded}, which did so through a planted pathway? Appendix Table~\ref{tab:auroc_all} scores them on flagging pathway-dominated interventions among all interventions, Table~\ref{tab:bootstrap} gives pooled AUROCs with bootstrap confidence intervals, and Table~\ref{tab:seeds} shows the spread over seeds. Figure~\ref{fig:bars} summarises Table~\ref{tab:auroc_success} by model.

\begin{table*}[!tp]
\centering
\footnotesize
\setlength{\tabcolsep}{2.2pt}
\caption{\textbf{Right answer, wrong mechanism.} AUROC restricted to \emph{successful} interventions (the counterfactual answer was produced); positives are those whose decision flips when the planted pathways are removed. Mean over configurations with $\geq$5 positives (\#cfg; at most 5 per GPT-2 row and 24 per row for the other models). Nearest-neighbour and local-PCA site distances are at or below chance in every row. $^\dagger$Measures used by \citet{grant2026divergent}.}
\label{tab:auroc_success}
\begin{tabular}{lll rr cccc c cccc}
\toprule
 & & & & & \multicolumn{4}{c}{\emph{site-level}} & \emph{downstream} & \multicolumn{4}{c}{\emph{downstream + causal (ours)}} \\
\cmidrule(lr){6-9}\cmidrule(lr){10-10}\cmidrule(lr){11-14}
Model & Task & Pathway & \#cfg & \#pos & kNN & Maha. & LPCA$^\dagger$ & Alg.~1$^\dagger$ & Down-OM & HPC-S & HPC-L & HPC-PCA & HPC-Loc \\
\midrule
GPT-2 & Gender & silent & 5 & 1099 & 0.38 & 0.69 & 0.38 & 0.65 & 0.69 & 0.83 & \textbf{0.83} & 0.83 & 0.82 \\
GPT-2 & Gender & hijack & 5 & 1393 & 0.40 & 0.65 & 0.40 & 0.57 & 0.67 & 0.57 & \textbf{0.87} & 0.85 & 0.85 \\
GPT-2 & Gender & distrib. & 5 & 832 & 0.35 & 0.73 & 0.35 & 0.69 & 0.68 & 0.62 & \textbf{0.87} & \textbf{0.87} & 0.86 \\
GPT-2 & Gender & in-range & 3 & 71 & 0.46 & 0.60 & 0.46 & 0.31 & 0.48 & 0.55 & 0.54 & \textbf{0.64} & 0.57 \\
GPT-2 & SVA & silent & 2 & 24 & 0.45 & 0.82 & 0.45 & 0.87 & 0.94 & \textbf{1.00} & \textbf{1.00} & \textbf{1.00} & 0.99 \\
GPT-2 & SVA & hijack & 2 & 14 & 0.47 & 0.81 & 0.47 & 0.86 & 0.93 & 0.77 & \textbf{0.99} & 0.93 & 0.95 \\
GPT-2 & SVA & distrib. & 4 & 27 & 0.39 & 0.77 & 0.39 & 0.68 & 0.92 & 0.73 & \textbf{0.96} & 0.93 & 0.89 \\
GPT-2 & SVA & in-range & 2 & 134 & 0.37 & 0.68 & 0.39 & 0.59 & 0.91 & 0.56 & \textbf{0.94} & 0.92 & 0.90 \\
\midrule
Pythia-1.4B & Gender & silent & 22 & 5543 & 0.36 & 0.69 & 0.36 & 0.67 & 0.64 & 0.84 & 0.86 & 0.87 & \textbf{0.87} \\
Pythia-1.4B & Gender & hijack & 24 & 5569 & 0.35 & 0.68 & 0.35 & 0.61 & 0.66 & 0.61 & 0.88 & 0.89 & \textbf{0.90} \\
Pythia-1.4B & Gender & distrib. & 23 & 3554 & 0.38 & 0.66 & 0.38 & 0.64 & 0.68 & 0.66 & 0.82 & 0.83 & \textbf{0.85} \\
Pythia-1.4B & Gender & in-range & 10 & 761 & 0.25 & 0.44 & 0.25 & 0.23 & 0.35 & 0.48 & 0.70 & 0.82 & \textbf{0.85} \\
Pythia-1.4B & SVA & silent & 11 & 912 & 0.53 & 0.50 & 0.54 & 0.83 & 0.76 & \textbf{0.93} & 0.92 & 0.88 & 0.91 \\
Pythia-1.4B & SVA & hijack & 11 & 1514 & 0.50 & 0.48 & 0.51 & 0.71 & 0.85 & 0.58 & \textbf{0.93} & 0.92 & \textbf{0.93} \\
Pythia-1.4B & SVA & distrib. & 11 & 1255 & 0.50 & 0.46 & 0.50 & 0.70 & 0.82 & 0.60 & \textbf{0.88} & 0.88 & 0.87 \\
Pythia-1.4B & SVA & in-range & 10 & 452 & 0.47 & 0.45 & 0.47 & 0.48 & 0.56 & 0.51 & 0.65 & 0.82 & \textbf{0.83} \\
\midrule
Pythia-6.9B & Gender & silent & 24 & 8491 & 0.36 & 0.61 & 0.35 & 0.68 & 0.57 & \textbf{0.82} & \textbf{0.83} & \textbf{0.83} & \textbf{0.83} \\
Pythia-6.9B & Gender & hijack & 24 & 8096 & 0.35 & 0.60 & 0.34 & 0.66 & 0.63 & 0.61 & 0.87 & \textbf{0.88} & \textbf{0.88} \\
Pythia-6.9B & Gender & distrib. & 24 & 6043 & 0.44 & 0.60 & 0.43 & 0.64 & 0.61 & 0.62 & 0.82 & \textbf{0.82} & \textbf{0.82} \\
Pythia-6.9B & Gender & in-range & 18 & 346 & 0.38 & 0.40 & 0.38 & 0.36 & 0.46 & 0.58 & 0.59 & 0.64 & \textbf{0.73} \\
Pythia-6.9B & SVA & silent & 19 & 3221 & 0.61 & 0.58 & 0.60 & 0.75 & 0.78 & \textbf{0.90} & \textbf{0.90} & 0.88 & 0.89 \\
Pythia-6.9B & SVA & hijack & 18 & 3520 & 0.60 & 0.58 & 0.59 & 0.72 & 0.82 & 0.57 & \textbf{0.91} & 0.90 & 0.90 \\
Pythia-6.9B & SVA & distrib. & 18 & 2958 & 0.62 & 0.54 & 0.62 & 0.76 & 0.84 & 0.62 & \textbf{0.92} & 0.90 & 0.91 \\
Pythia-6.9B & SVA & in-range & 17 & 836 & 0.54 & 0.50 & 0.54 & 0.46 & 0.56 & 0.50 & 0.65 & \textbf{0.77} & 0.75 \\
\midrule
Qwen2.5-7B & Gender & silent & 21 & 3634 & 0.21 & 0.32 & 0.21 & 0.71 & 0.69 & \textbf{0.90} & 0.89 & 0.89 & 0.88 \\
Qwen2.5-7B & Gender & hijack & 21 & 3363 & 0.25 & 0.36 & 0.26 & 0.62 & 0.79 & 0.62 & \textbf{0.94} & 0.91 & 0.91 \\
Qwen2.5-7B & Gender & distrib. & 21 & 2911 & 0.31 & 0.38 & 0.31 & 0.86 & 0.90 & 0.60 & \textbf{0.94} & \textbf{0.94} & 0.93 \\
Qwen2.5-7B & Gender & in-range & 6 & 509 & 0.32 & 0.50 & 0.32 & 0.31 & 0.46 & 0.50 & 0.39 & \textbf{0.55} & 0.48 \\
Qwen2.5-7B & SVA & silent & 17 & 1705 & 0.22 & 0.28 & 0.22 & 0.85 & 0.86 & \textbf{0.97} & 0.96 & 0.94 & 0.93 \\
Qwen2.5-7B & SVA & hijack & 22 & 1758 & 0.26 & 0.30 & 0.26 & 0.77 & 0.91 & 0.54 & \textbf{0.96} & 0.93 & 0.90 \\
Qwen2.5-7B & SVA & distrib. & 13 & 1173 & 0.23 & 0.32 & 0.23 & 0.82 & 0.92 & 0.51 & \textbf{0.96} & 0.95 & 0.94 \\
Qwen2.5-7B & SVA & in-range & 12 & 915 & 0.20 & 0.30 & 0.20 & 0.54 & 0.58 & 0.50 & 0.63 & \textbf{0.85} & \textbf{0.85} \\
\bottomrule
\end{tabular}
\end{table*}

\paragraph{Site-level distance is uninformative, even misleading, where it matters.} On the gender task, kNN and LPCA fall below 0.5 for every model and every pathway family (0.21--0.46): wrong-mechanism successes sit \emph{closer} to the natural site manifold than genuine ones. The same holds for SVA on GPT-2 and Qwen2.5-7B (0.20--0.47). On SVA in the two Pythia models they are near chance (0.47--0.62), with large variation across configurations. A plausible reason is that confirmatory pathways let small, typical-looking perturbations tip the decision, while genuine successes often require larger moves. Mahalanobis distance and Algorithm~1 are better on some rows but inconsistent: Algorithm~1 ranges from 0.23 to 0.87 and is below chance on every in-range row except the two SVA rows of GPT-2 and Qwen2.5-7B (0.59 and 0.54).

\paragraph{Looking downstream helps, and the causal test helps most.} Over all interventions, Down-OM is strong (Table~\ref{tab:auroc_all}, 0.86--0.99 on unit-visible families). Among successful interventions it drops (0.57--0.94), while HPC-L stays at 0.82--1.00 for the silent, hijack and distributed families on every model. The step of clamping and measuring the lost margin separates ``unusual'' from ``used for the decision''. The ordering is stable across seeds: for every model, task and unit-visible family, the per-seed mean of HPC-L exceeds that of LPCA by at least 0.26 (Table~\ref{tab:seeds}).

\paragraph{What \hpc{} clamps matters, and in-range pathways defeat range clamping.} HPC-S, which only silences units that are never active, fails when a pathway hijacks neurons that are active naturally (0.51--0.77 on hijack and distributed). Range clamping (HPC-L) handles these cases. It does not handle \emph{in-range} pathways: among successful interventions HPC-L scores 0.59--0.70 on the Pythia models and Qwen2.5-7B SVA, and 0.39 on Qwen2.5-7B gender. On GPT-2 it is high on SVA (0.94) but at chance on gender (0.54).

\paragraph{Projecting onto the natural manifold partly closes the gap, but not uniformly.} The projection variants do better on in-range pathways at scale. Among successful interventions, HPC-Loc reaches 0.83--0.85 on Pythia-1.4B, 0.73--0.75 on Pythia-6.9B and 0.85 on Qwen2.5-7B SVA (HPC-PCA: 0.64--0.85), against 0.57 on GPT-2 gender and 0.48 on Qwen2.5-7B gender. Over all interventions, HPC-Loc reaches 0.85--0.91 on in-range pathways in the Pythia models (Table~\ref{tab:auroc_all}). These are the first signs that a combination-level test can see in-range pathways. However, the gains are model- and task-dependent, the projection variants are not uniformly better on the unit-visible families, and on Qwen2.5-7B gender no detector we tested is reliably above chance among successful interventions. We therefore still regard in-range pathways as the open problem the benchmark exposes. Note that few Qwen2.5-7B gender configurations have enough wrong-mechanism successes to be scored (6 of 24), because the model's natural success rate at the subject token is low (\S\ref{sec:res-natural}).

\subsection{Optimised interventions seek hidden pathways}
\label{sec:res-illusion}

Table~\ref{tab:illusion_scale} reports what DAS does when hidden pathways exist, for every model.

\begin{table}[!htbp]
\centering
\footnotesize
\setlength{\tabcolsep}{2.6pt}
\caption{\textbf{Optimised interventions find hidden pathways, at every scale.} Success rate (succ., \%) and, among successes, the pathway-dominated share (path., \%), averaged over configurations. D: DAS trained on the unmodified model, applied with pathways present. G: DAS trained on the model with pathways. H: restricted DAS (top-32 site components and the D direction excluded). The path.\ column for H averages only over configurations where H succeeded at least once (last column: configurations with H successes / all configurations). GPT-2: seed 0; other models: seeds 0--2.}
\label{tab:illusion_scale}
\begin{tabular}{lll r cc cc cc r}
\toprule
 & & & & \multicolumn{2}{c}{D: clean DAS} & \multicolumn{2}{c}{G: DAS} & \multicolumn{2}{c}{H: restricted} & \\
\cmidrule(lr){5-6}\cmidrule(lr){7-8}\cmidrule(lr){9-10}
Model & Task & Pathway & \#cfg & succ. & path. & succ. & path. & succ. & path. & H cfg \\
\midrule
GPT-2 & Gender & silent & 5 & 39.3 & 63.0 & 91.2 & 90.2 & 13.5 & 100.0 & 5/5 \\
GPT-2 & Gender & hijack & 5 & 46.6 & 77.3 & 90.7 & 90.0 & 26.1 & 100.0 & 5/5 \\
GPT-2 & Gender & distrib. & 5 & 42.3 & 55.7 & 87.3 & 95.1 & 12.4 & 100.0 & 5/5 \\
GPT-2 & Gender & in-range & 5 & 70.0 & 5.8 & 68.1 & 16.5 & 0.9 & 75.0 & 3/5 \\
GPT-2 & SVA & silent & 5 & 47.1 & 2.1 & 67.2 & 1.0 & 3.1 & 45.8 & 5/5 \\
GPT-2 & SVA & hijack & 5 & 57.5 & 3.4 & 69.2 & 0.2 & 1.9 & 45.0 & 5/5 \\
GPT-2 & SVA & distrib. & 5 & 50.8 & 2.4 & 62.3 & 7.1 & 2.8 & 80.0 & 5/5 \\
GPT-2 & SVA & in-range & 5 & 67.3 & 0.9 & 65.6 & 6.7 & 11.0 & 66.7 & 5/5 \\
\midrule
Pythia-1.4B & Gender & silent & 24 & 42.7 & 52.7 & 78.1 & 80.5 & 19.7 & 84.8 & 23/24 \\
Pythia-1.4B & Gender & hijack & 24 & 44.1 & 46.5 & 77.4 & 85.9 & 20.4 & 88.9 & 24/24 \\
Pythia-1.4B & Gender & distrib. & 24 & 46.5 & 39.0 & 70.1 & 90.6 & 15.7 & 76.2 & 23/24 \\
Pythia-1.4B & Gender & in-range & 24 & 62.2 & 3.0 & 61.7 & 15.9 & 1.4 & 30.0 & 15/24 \\
Pythia-1.4B & SVA & silent & 24 & 69.8 & 3.0 & 73.7 & 17.1 & 16.2 & 20.5 & 24/24 \\
Pythia-1.4B & SVA & hijack & 24 & 71.0 & 2.1 & 75.5 & 25.5 & 23.9 & 24.9 & 24/24 \\
Pythia-1.4B & SVA & distrib. & 24 & 71.3 & 2.0 & 72.4 & 21.7 & 22.1 & 24.7 & 24/24 \\
Pythia-1.4B & SVA & in-range & 24 & 75.7 & 1.7 & 73.7 & 17.7 & 9.8 & 15.7 & 24/24 \\
\midrule
Pythia-6.9B & Gender & silent & 24 & 55.5 & 62.5 & 84.5 & 96.6 & 17.1 & 97.7 & 24/24 \\
Pythia-6.9B & Gender & hijack & 24 & 51.6 & 59.2 & 76.3 & 95.8 & 18.1 & 91.7 & 24/24 \\
Pythia-6.9B & Gender & distrib. & 24 & 51.1 & 46.8 & 70.2 & 86.3 & 14.5 & 91.7 & 24/24 \\
Pythia-6.9B & Gender & in-range & 24 & 55.1 & 1.6 & 54.9 & 5.7 & 2.2 & 60.5 & 19/24 \\
Pythia-6.9B & SVA & silent & 24 & 57.4 & 20.3 & 69.2 & 47.1 & 21.6 & 50.0 & 24/24 \\
Pythia-6.9B & SVA & hijack & 24 & 58.2 & 24.3 & 67.0 & 45.6 & 24.3 & 50.7 & 24/24 \\
Pythia-6.9B & SVA & distrib. & 24 & 61.2 & 13.4 & 71.5 & 46.3 & 21.9 & 50.3 & 24/24 \\
Pythia-6.9B & SVA & in-range & 24 & 65.6 & 4.6 & 64.8 & 13.9 & 11.7 & 42.1 & 24/24 \\
\midrule
Qwen2.5-7B & Gender & silent & 24 & 14.9 & 60.4 & 39.2 & 99.3 & 9.5 & 100.0 & 6/24 \\
Qwen2.5-7B & Gender & hijack & 24 & 14.7 & 55.4 & 33.3 & 99.6 & 9.9 & 99.8 & 6/24 \\
Qwen2.5-7B & Gender & distrib. & 24 & 16.5 & 30.7 & 36.8 & 94.6 & 7.6 & 100.0 & 7/24 \\
Qwen2.5-7B & Gender & in-range & 24 & 11.9 & 5.5 & 15.8 & 24.9 & 1.1 & 100.0 & 3/24 \\
Qwen2.5-7B & SVA & silent & 24 & 33.9 & 9.7 & 38.4 & 64.4 & 7.6 & 100.0 & 6/24 \\
Qwen2.5-7B & SVA & hijack & 24 & 34.5 & 12.9 & 31.9 & 75.9 & 9.0 & 100.0 & 6/24 \\
Qwen2.5-7B & SVA & distrib. & 24 & 37.7 & 4.1 & 30.0 & 46.9 & 9.7 & 99.9 & 7/24 \\
Qwen2.5-7B & SVA & in-range & 24 & 36.2 & 4.3 & 32.6 & 38.2 & 6.7 & 100.0 & 6/24 \\
\bottomrule
\end{tabular}
\end{table}

\paragraph{Unrestricted DAS.} On the gender task, DAS trained on a model with silent, hijack or distributed pathways (G) routes 81--100\% of its successes through them on every model: 90--95\% on GPT-2, 81--91\% on Pythia-1.4B, 86--97\% on Pythia-6.9B and 95--100\% on Qwen2.5-7B. Gradient descent prefers the planted shortcut to the model's own mechanism. On SVA, where the natural signal is strong, GPT-2 DAS largely ignores the pathways (0.2--7\%), but the share grows with the larger models: 17--26\% on Pythia-1.4B, 46--47\% on Pythia-6.9B and 47--76\% on Qwen2.5-7B. Even DAS trained on the unmodified model (D) produces pathway-dominated successes when applied with overshoot: 39--63\% of its gender successes in the Pythia models and 31--60\% in Qwen2.5-7B.

\paragraph{Restricted DAS.} With the genuine causal subspace excluded (H), DAS still reports success on gender for the three unit-visible families: 12--26\% of examples on GPT-2, 16--20\% on Pythia-1.4B, 14--18\% on Pythia-6.9B and 8--10\% on Qwen2.5-7B. Of these successes, 100\% are pathway-dominated on GPT-2, 76--89\% on Pythia-1.4B, 92--98\% on Pythia-6.9B and essentially all on Qwen2.5-7B, where only 6--7 of 24 configurations produce any H success. On SVA, restricted DAS succeeds more often in the Pythia models (16--24\%), but only 20--51\% of those successes are pathway-dominated. This is consistent with the SVA signal extending beyond the top 32 components. In-range pathways are rarely exploited (H success $\leq2.2\%$ on gender). This reproduces the interpretability illusion of \citet{makelov2024subspace} as a controlled measurement at four scales: an optimiser searching for an intervention direction finds hidden pathways when they are easier to use than the natural mechanism, and a high interchange-intervention accuracy is then no evidence about that mechanism.

\paragraph{Mitigation (GPT-2).} We add to the DAS objective a penalty equal to $\lambda$ times the mean relative distance of downstream MLP activations from the natural class-conditional PCA subspace of the counterfactual class (Down-OM as a differentiable loss, $\lambda=5$). Table~\ref{tab:illusion} shows two effects. First, for unrestricted DAS on the gender task, the matched comparison within the mitigation runs shows that the penalty lowers the success rate from 94--98\% (G$'$) to 87--89\% (J). It cuts the share of successes that are pathway-dominated from 98.5--100\% to 4.5--4.7\%: the penalised direction reaches the counterfactual answer through the model's own computation. Second, for restricted DAS, the penalty lowers the success rate on gender (from H$'$ to I: 52.5\% to 41.0\% for silent, 66.0\% to 17.0\% for hijack, 34.5\% to 30.0\% for distributed). The successes that remain are still 92--100\% pathway-dominated. When the genuine subspace is unavailable, the penalty thus reduces but does not remove false successes, and restricted-subspace results should not be trusted even with the penalty. On SVA, penalised restricted DAS almost never succeeds (0--1\%). We have evaluated the penalty only on GPT-2.

\begin{table*}[!tp]
\centering
\small
\setlength{\tabcolsep}{3.2pt}
\caption{\textbf{Mitigation on GPT-2: a downstream on-manifold penalty redirects unrestricted DAS.} For each DAS variant: success rate (succ., \%) and, among its successes, the share that is pathway-dominated (path., \%). G: DAS trained on the model with planted pathways. H: the same, restricted to the complement of the top-32 principal components of the site and of the clean-model DAS direction (illusion probe). Main runs: all sites, seed 0. Mitigation runs (GPT-2, $L{=}2$, $T{=}4$, 900 interventions; gender: seeds 0--1, SVA: seed 0) retrain the same variants without (G$'$, H$'$) and with (J, I) the downstream on-manifold penalty ($\lambda{=}5$).}
\label{tab:illusion}
\begin{tabular}{ll cc cc | cc cc | cc cc}
\toprule
 & & \multicolumn{4}{c|}{\emph{main runs}} & \multicolumn{8}{c}{\emph{mitigation runs}} \\
 & & \multicolumn{2}{c}{G: DAS} & \multicolumn{2}{c|}{H: restricted} & \multicolumn{2}{c}{G$'$: DAS} & \multicolumn{2}{c|}{J: DAS+pen.} & \multicolumn{2}{c}{H$'$: restr.} & \multicolumn{2}{c}{I: restr.+pen.} \\
Task & Pathway & succ. & path. & succ. & path. & succ. & path. & succ. & path. & succ. & path. & succ. & path. \\
\midrule
Gender & silent & 91.2 & 90.2 & 13.5 & 100.0 & 98.0 & 98.5 & 87.0 & 4.5 & 52.5 & 100.0 & 41.0 & 100.0 \\
Gender & hijack & 90.7 & 90.0 & 26.1 & 100.0 & 94.0 & 100.0 & 88.0 & 4.7 & 66.0 & 100.0 & 17.0 & 100.0 \\
Gender & distrib. & 87.3 & 95.1 & 12.4 & 100.0 & 96.0 & 100.0 & 89.0 & 4.5 & 34.5 & 100.0 & 30.0 & 92.3 \\
Gender & in-range & 68.1 & 16.5 & 0.9 & 75.0 & 90.0 & 0.0 & 88.5 & 0.0 & 3.0 & 40.0 & 6.5 & 0.0 \\
SVA & silent & 67.2 & 1.0 & 3.1 & 45.8 & 68.0 & 0.0 & 78.0 & 0.0 & 3.0 & 100.0 & 0.0 & -- \\
SVA & hijack & 69.2 & 0.2 & 1.9 & 45.0 & 78.0 & 0.0 & 78.0 & 0.0 & 3.0 & 100.0 & 0.0 & -- \\
SVA & distrib. & 62.3 & 7.1 & 2.8 & 80.0 & 86.0 & 2.3 & 83.0 & 0.0 & 15.0 & 100.0 & 1.0 & 100.0 \\
SVA & in-range & 65.6 & 6.7 & 11.0 & 66.7 & 85.0 & 0.0 & 84.0 & 0.0 & 0.0 & -- & 0.0 & -- \\
\bottomrule
\end{tabular}
\end{table*}

\subsection{Unmodified models}
\label{sec:res-natural}

Do real interventions in unmodified models rely on out-of-regime downstream activity? Table~\ref{tab:natural} reports the fraction of successful interventions with HPC-L $>1$, i.e.\ where out-of-regime activity supplies more than a typical natural margin. Table~\ref{tab:ksweep} varies the number $k$ of principal components excluded from restricted DAS, and Table~\ref{tab:posctl} gives the positive control.

\begin{table}[!htbp]
\centering
\footnotesize
\setlength{\tabcolsep}{2.8pt}
\caption{\textbf{Unmodified models.} Percentage of \emph{successful} interventions whose decision relies on out-of-regime downstream activity (HPC-L $>1$, i.e.\ more than one natural median margin), averaged over configurations with $\geq$5 successes of that type; mean Spearman $\rho$ between site divergence (LPCA) and HPC-L; success rate (\%) of restricted DAS (H, top 32 components excluded). Mean-diff.\ pools both mean-difference groups (C and the steering group).}
\label{tab:natural}
\begin{tabular}{ll rr cccc c c}
\toprule
Model & Task & \#cfg & \#succ. & full patch & mean-diff. & DAS & DAS+rand. & $\rho$ & H succ. \\
\midrule
GPT-2 & Gender & 4 & 2720 & 0.0 & 0.0 & 0.8 & 0.0 & -0.15 & 1.3 \\
GPT-2 & SVA & 4 & 2809 & 0.0 & 0.0 & 0.0 & 0.0 & -0.13 & 0.5 \\
GPT-2 & IOI & 4 & 2174 & 0.0 & 0.0 & 0.3 & 0.0 & 0.21 & 0.0 \\
\midrule
Pythia-1.4B & Gender & 12 & 8030 & 0.0 & 0.1 & 0.0 & 0.0 & -0.00 & 0.8 \\
Pythia-1.4B & SVA & 12 & 8344 & 0.0 & 0.0 & 0.0 & 0.0 & 0.02 & 8.7 \\
\midrule
Pythia-6.9B & Gender & 12 & 6732 & 0.0 & 0.2 & 0.2 & 0.0 & 0.18 & 1.1 \\
Pythia-6.9B & SVA & 12 & 7617 & 0.0 & 0.0 & 0.0 & 0.0 & 0.04 & 8.5 \\
Pythia-6.9B & IOI & 12 & 4335 & 0.0 & 0.0 & 0.0 & 0.0 & 0.04 & 2.4 \\
\midrule
Qwen2.5-7B & Gender & 12 & 1053 & 0.0 & 0.0 & 0.0 & 0.0 & -0.12 & 0.1 \\
Qwen2.5-7B & SVA & 12 & 2844 & 0.0 & 0.0 & 0.0 & 0.0 & -0.11 & 2.1 \\
Qwen2.5-7B & IOI & 12 & 5247 & 0.0 & 0.0 & 0.0 & 0.0 & 0.01 & 0.5 \\
\bottomrule
\end{tabular}
\end{table}

\paragraph{Standard interventions.} For full patching, mean-difference patching, DAS and DAS mixed with random directions, the fraction is at most 0.8\% in every model and task, and exactly zero in Qwen2.5-7B. Site divergence is only weakly correlated with HPC-L (mean $\rho$ between $-0.15$ and $0.21$). Success rates differ strongly between models: in Qwen2.5-7B, interventions at the subject token rarely change the gender answer (Table~\ref{tab:natural}, 1{,}053 successes over 12 configurations), so the conditional rate there rests on few examples.

\begin{table}[!htbp]
\centering
\footnotesize
\setlength{\tabcolsep}{3pt}
\caption{\textbf{Restricted DAS in unmodified models}, excluding the top $k$ principal components of the site and the clean DAS direction. Success rate (\%, mean and maximum over configurations), number of successes, and how many of them have HPC-L $>1$ or HPC-S $>1$. Rows with $k{=}32$ for gender and SVA are the no-pathway configurations of the main benchmark.}
\label{tab:ksweep}
\begin{tabular}{ll r r rr r rr r}
\toprule
 & & & & \multicolumn{2}{c}{H succ.\ (\%)} & & \multicolumn{2}{c}{HPC-L$>$1} & HPC-S \\ \cmidrule(lr){5-6}\cmidrule(lr){8-9}
Model & Task & $k$ & \#cfg & mean & max & \#succ. & \# & \% & $>$1 \\
\midrule
GPT-2 & Gender & 32 & 4 & 1.3 & 2.8 & 11 & 0 & 0.0 & 0 \\
\midrule
GPT-2 & SVA & 32 & 4 & 0.5 & 0.9 & 4 & 0 & 0.0 & 0 \\
\midrule
GPT-2 & IOI & 4 & 8 & 12.3 & 31.3 & 210 & 0 & 0.0 & 0 \\
GPT-2 & IOI & 32 & 4 & 0.0 & 0.0 & 0 & 0 & -- & 0 \\
\midrule
Pythia-1.4B & Gender & 32 & 12 & 0.8 & 3.7 & 21 & 0 & 0.0 & 0 \\
\midrule
Pythia-1.4B & SVA & 32 & 12 & 8.7 & 16.8 & 223 & 0 & 0.0 & 0 \\
\midrule
Pythia-6.9B & Gender & 1 & 12 & 30.4 & 57.0 & 780 & 0 & 0.0 & 0 \\
Pythia-6.9B & Gender & 4 & 12 & 27.2 & 54.7 & 698 & 0 & 0.0 & 0 \\
Pythia-6.9B & Gender & 32 & 12 & 1.1 & 4.2 & 29 & 0 & 0.0 & 0 \\
\midrule
Pythia-6.9B & SVA & 1 & 12 & 32.7 & 56.5 & 841 & 0 & 0.0 & 0 \\
Pythia-6.9B & SVA & 4 & 12 & 18.7 & 45.3 & 479 & 0 & 0.0 & 0 \\
Pythia-6.9B & SVA & 32 & 12 & 8.5 & 16.8 & 218 & 0 & 0.0 & 0 \\
\midrule
Pythia-6.9B & IOI & 1 & 12 & 23.8 & 33.6 & 611 & 0 & 0.0 & 0 \\
Pythia-6.9B & IOI & 4 & 12 & 8.8 & 16.8 & 226 & 0 & 0.0 & 0 \\
Pythia-6.9B & IOI & 32 & 12 & 2.4 & 9.8 & 61 & 0 & 0.0 & 0 \\
\midrule
Qwen2.5-7B & Gender & 1 & 12 & 2.8 & 15.0 & 71 & 10 & 14.1 & 0 \\
Qwen2.5-7B & Gender & 4 & 12 & 2.7 & 15.9 & 69 & 8 & 11.6 & 0 \\
Qwen2.5-7B & Gender & 32 & 12 & 0.1 & 0.5 & 3 & 3 & 100.0 & 0 \\
\midrule
Qwen2.5-7B & SVA & 1 & 12 & 6.0 & 28.0 & 154 & 49 & 31.8 & 0 \\
Qwen2.5-7B & SVA & 4 & 12 & 6.0 & 28.0 & 155 & 47 & 30.3 & 0 \\
Qwen2.5-7B & SVA & 32 & 12 & 2.1 & 11.2 & 55 & 2 & 3.6 & 0 \\
\midrule
Qwen2.5-7B & IOI & 1 & 12 & 8.0 & 27.1 & 206 & 0 & 0.0 & 0 \\
Qwen2.5-7B & IOI & 4 & 12 & 5.5 & 12.1 & 140 & 0 & 0.0 & 0 \\
Qwen2.5-7B & IOI & 32 & 12 & 0.5 & 3.3 & 14 & 0 & 0.0 & 0 \\
\bottomrule
\end{tabular}
\end{table}

\paragraph{Restricted DAS without planted pathways.} With the top 32 components excluded, restricted DAS rarely succeeds (at most 8.7\% on average). Excluding fewer components makes it succeed much more often. In Pythia-6.9B, with $k{=}1$ it succeeds on 24--33\% of examples on average (up to 57\% in one configuration), yet none of its 2{,}232 successes across the three tasks shows HPC-L $>1$. The same holds for GPT-2 IOI with $k{=}4$ (\IOIkFourSucc\% on average, up to \IOIkFourMax\%; \IOIkFourHPC{} of \IOIkFourN{} successes) and for Qwen2.5-7B IOI. This is consistent with the genuine signal extending beyond a few principal directions~\citep{wu2024reply}, rather than with a hidden pathway.

\paragraph{Candidate wrong-mechanism successes in Qwen2.5-7B.} The exception is Qwen2.5-7B on SVA and gender. With $k\in\{1,4\}$, 30--32\% of restricted-DAS successes on SVA (49 of 154 and 47 of 155) and 12--14\% on gender (10 of 71 and 8 of 69) have HPC-L $>1$, with values up to about 6 natural margins. All of them occur at the final-token site ($L{=}20$), where full patches and mean-difference patching show no such cases and DAS shows 2 in 808 successes. None exceeds 1 under HPC-S, so they involve naturally active neurons driven outside their range, not dormant units. Compared with full patches at the same site, these interventions have a larger decision margin, a 2.5--7 times larger site LPCA distance and about twice the Down-OM score; about two thirds of them also exceed 1 under HPC-PCA and about half under HPC-Loc. This is the signature that the unit-visible pathway families produce in the benchmark. We stress that it is a \emph{candidate} signature: without planted ground truth we cannot confirm that these decisions are produced by a mechanism the model does not use, and clamping can over-estimate dependence when later layers compensate. They are, however, exactly the interventions a researcher would want to inspect before using restricted-subspace success as evidence.

\begin{table}[!htbp]
\centering
\footnotesize
\setlength{\tabcolsep}{4pt}
\caption{\textbf{Positive control for Table~\ref{tab:natural}.} The same statistic (HPC-L $>1$, \%) among wrong-mechanism successes in models with planted pathways, all intervention types pooled; unit-visible families (silent, hijack, distributed) vs.\ in-range. \#: number of wrong-mechanism successes.}
\label{tab:posctl}
\begin{tabular}{ll rc rc}
\toprule
 & & \multicolumn{2}{c}{unit-visible} & \multicolumn{2}{c}{in-range} \\ \cmidrule(lr){3-4}\cmidrule(lr){5-6}
Model & Task & \# & HPC-L$>$1 & \# & HPC-L$>$1 \\
\midrule
GPT-2 & Gender & 3324 & 90.1 & 71 & 0.0 \\
GPT-2 & SVA & 65 & 80.0 & 134 & 1.5 \\
Pythia-1.4B & Gender & 14666 & 93.2 & 761 & 2.6 \\
Pythia-1.4B & SVA & 3681 & 92.3 & 452 & 0.0 \\
Pythia-6.9B & Gender & 22630 & 94.0 & 346 & 0.3 \\
Pythia-6.9B & SVA & 9699 & 89.7 & 836 & 0.0 \\
Qwen2.5-7B & Gender & 9908 & 82.3 & 509 & 0.0 \\
Qwen2.5-7B & SVA & 4636 & 86.8 & 915 & 0.7 \\
\bottomrule
\end{tabular}
\end{table}

\paragraph{What these checks can see.} The positive control bounds what Tables~\ref{tab:natural} and~\ref{tab:ksweep} tell us. In models with planted pathways, the same statistic catches 80--94\% of wrong-mechanism successes for the unit-visible families on every model, but at most 2.6\% for in-range pathways. The honest conclusion is therefore limited. Across four models from 124M to 7B parameters, standard interventions on these tasks do not rely on \emph{unit-level} out-of-regime activity; restricted DAS in Qwen2.5-7B sometimes appears to. Reliance on in-range combinations would go undetected by HPC-L.

\subsection{Robustness}
\label{sec:robust}

\paragraph{Seeds.} For the three larger models, every configuration was run with three seeds, which change the prompt split, the planted directions and the DAS initialisation. Per-seed means move by a few hundredths for HPC-L on unit-visible families and by more on in-range rows and for site-level measures (Table~\ref{tab:seeds}); no qualitative conclusion changes between seeds. For GPT-2, seeds 1--2 were run separately on the main sites at a larger budget (Appendix Table~\ref{tab:ascend_gpt2}); they reproduce the seed-0 pattern on gender: LPCA 0.31--0.39 and HPC-L 0.86--0.93 on unit-visible families, HPC-L 0.58 on in-range, and 92--96\% of DAS (G) successes pathway-dominated. The completed SVA configurations have too few wrong-mechanism successes to be scored.

\paragraph{Hardware.} Re-running one GPT-2 configuration with the same code on a different GPU changed several AUROCs by up to 0.21 among successful interventions, together with the number of wrong-mechanism successes (Appendix Table~\ref{tab:p0}). Scores on the pathway-dominated label over all interventions agree closely for the HPC variants. Same-seed runs are therefore not bit-reproducible across hardware, because SVD bases and DAS optimisation differ; we report per-configuration results only as averages over many configurations.

\paragraph{Instruction tuning (exploratory).} A small-budget comparison of Qwen2.5-1.5B and its Instruct variant (Appendix Table~\ref{tab:qwen_preview}) shows the same pattern for hijack pathways in both: kNN below chance (0.37 base, 0.43 Instruct) and HPC-L high (0.85 and 0.92). Without planted pathways, 0 of 591 and 1 of 599 successful interventions have HPC-L $>1$. This is a preliminary check with raw-completion prompts and is not evidence about chat-formatted use.

\section{Discussion and limitations}

\paragraph{Planted pathways are constructions.} They show what a detector can and cannot see when a hidden pathway exists, not how often such pathways arise naturally. Three of our four families are, by design, visible to unit-level range tests. They are sanity checks, and near-perfect scores on them should not be read as evidence of real-world reliability. The in-range family is the informative one. Range clamping fails on it at every scale except GPT-2 SVA, and the projection variants help on some models and tasks but not others.

\paragraph{Downstream self-repair.} Clamping downstream units can trigger compensation by later components~\citep{mcgrath2023hydra}. This makes $\LD_\Pi$ an imperfect estimate of the margin that would be lost, which matters most for the natural-model candidates of \S\ref{sec:res-natural}.

\paragraph{Statistics and constants.} GPT-2 main configurations use a single seed; the larger models use three. The tables report means over configurations, which share prompts and baselines and are not independent replicates; per-configuration standard deviations are not confidence intervals. Several constants are fixed rather than tuned: the pathway gain, the ten-times in-range gain, the quantiles defining a regime, and the HPC-L $>1$ threshold. Natural regimes are estimated on the same templated prompts that are intervened on; a held-out natural distribution would be a stricter reference.

\paragraph{Scope.} The largest model has 7B parameters. For Pythia-6.9B and Qwen2.5-7B, downstream detectors analyse every second MLP layer after the site (always including the pathway layer $T$), and models run in bfloat16; both choices weaken the natural-model check relative to full coverage. \hpc{} needs a class-conditional decision to define ``natural runs with the same output''. It clamps MLP neurons at the final position only, so pathways mediated by attention or located at other positions are not covered. The mitigation has been evaluated only on GPT-2.

\paragraph{Claim dependence.} Following \citet{grant2026divergent}, harm is relative to a claim. \hpc{} targets claims about the natural mechanism. For other claims, e.g.\ that a subspace is \emph{sufficient} to control behaviour, out-of-regime computation may be acceptable.

\section{Conclusion}
Whether an intervention gives the right answer for the right reason cannot be read off the intervened representation. It has to be tested downstream, against what the model does on natural inputs. Planting hidden pathways in pretrained models turns this into a measurable question. From 124M to 7B parameters, site-level distances fail on that benchmark, and a simple downstream causal test succeeds whenever the pathway is visible at the level of individual units. Optimised interventions exploit hidden pathways when these are easier to use than the natural mechanism, and in an unmodified 7B model restricted DAS produces successes with the same downstream signature. Pathways hidden in in-range combinations of units are only partly detected by projecting onto the natural manifold, and not at all on one model--task pair. We offer the benchmark as a target for detectors that can find them.

\FloatBarrier
\bibliographystyle{plainnat}
\bibliography{refs}

@inproceedings{grant2026divergent,
  title     = {Addressing Divergent Representations from Causal Interventions on Neural Networks},
  author    = {Grant, Satchel and Han, Simon Jerome and Tartaglini, Alexa R. and Potts, Christopher},
  booktitle = {International Conference on Learning Representations (ICLR)},
  year      = {2026},
  note      = {arXiv:2511.04638}
}

@inproceedings{makelov2024subspace,
  title     = {Is This the Subspace You Are Looking for? {A}n Interpretability Illusion for Subspace Activation Patching},
  author    = {Makelov, Aleksandar and Lange, Georg and Nanda, Neel},
  booktitle = {International Conference on Learning Representations (ICLR)},
  year      = {2024},
  note      = {arXiv:2311.17030}
}

@inproceedings{luo2026glp,
  title     = {Learning a Generative Meta-Model of {LLM} Activations},
  author    = {Luo, Grace and Feng, Jiahai and Darrell, Trevor and Radford, Alec and Steinhardt, Jacob},
  booktitle = {International Conference on Machine Learning (ICML)},
  year      = {2026},
  note      = {arXiv:2602.06964}
}

@article{vaidyanathan2026curse,
  title   = {The Curse of Multiple Mediators: Hidden Interaction Effects in Activation Patching},
  author  = {Vaidyanathan, Sankaran and Arbour, David and Mueller, Aaron and Niekum, Scott and Jensen, David},
  journal = {arXiv preprint arXiv:2606.27510},
  year    = {2026}
}

@inproceedings{sutter2025nonlinear,
  title     = {The Non-Linear Representation Dilemma: Is Causal Abstraction Enough for Mechanistic Interpretability?},
  author    = {Sutter, Denis and Minder, Julian and Hofmann, Thomas and Pimentel, Tiago},
  booktitle = {Advances in Neural Information Processing Systems (NeurIPS)},
  year      = {2025}
}

@inproceedings{geiger2021causal,
  title     = {Causal Abstractions of Neural Networks},
  author    = {Geiger, Atticus and Lu, Hanson and Icard, Thomas and Potts, Christopher},
  booktitle = {Advances in Neural Information Processing Systems (NeurIPS)},
  year      = {2021}
}

@inproceedings{geiger2024das,
  title     = {Finding Alignments Between Interpretable Causal Variables and Distributed Neural Representations},
  author    = {Geiger, Atticus and Wu, Zhengxuan and Potts, Christopher and Icard, Thomas and Goodman, Noah},
  booktitle = {Causal Learning and Reasoning (CLeaR)},
  year      = {2024}
}

@inproceedings{wu2023boundless,
  title     = {Interpretability at Scale: Identifying Causal Mechanisms in {A}lpaca},
  author    = {Wu, Zhengxuan and Geiger, Atticus and Icard, Thomas and Potts, Christopher and Goodman, Noah},
  booktitle = {Advances in Neural Information Processing Systems (NeurIPS)},
  year      = {2023}
}

@inproceedings{vig2020causal,
  title     = {Investigating Gender Bias in Language Models Using Causal Mediation Analysis},
  author    = {Vig, Jesse and Gehrmann, Sebastian and Belinkov, Yonatan and Qian, Sharon and Nevo, Daniel and Singer, Yaron and Shieber, Stuart},
  booktitle = {Advances in Neural Information Processing Systems (NeurIPS)},
  year      = {2020}
}

@inproceedings{meng2022rome,
  title     = {Locating and Editing Factual Associations in {GPT}},
  author    = {Meng, Kevin and Bau, David and Andonian, Alex and Belinkov, Yonatan},
  booktitle = {Advances in Neural Information Processing Systems (NeurIPS)},
  year      = {2022}
}

@inproceedings{wang2023ioi,
  title     = {Interpretability in the Wild: a Circuit for Indirect Object Identification in {GPT-2} Small},
  author    = {Wang, Kevin and Variengien, Alexandre and Conmy, Arthur and Shlegeris, Buck and Steinhardt, Jacob},
  booktitle = {International Conference on Learning Representations (ICLR)},
  year      = {2023}
}

@inproceedings{zhang2024patching,
  title     = {Towards Best Practices of Activation Patching in Language Models: Metrics and Methods},
  author    = {Zhang, Fred and Nanda, Neel},
  booktitle = {International Conference on Learning Representations (ICLR)},
  year      = {2024}
}

@article{heimersheim2024patching,
  title   = {How to Use and Interpret Activation Patching},
  author  = {Heimersheim, Stefan and Nanda, Neel},
  journal = {arXiv preprint arXiv:2404.15255},
  year    = {2024}
}

@inproceedings{conmy2023acdc,
  title     = {Towards Automated Circuit Discovery for Mechanistic Interpretability},
  author    = {Conmy, Arthur and Mavor-Parker, Augustine N. and Lynch, Aengus and Heimersheim, Stefan and Garriga-Alonso, Adri{\`a}},
  booktitle = {Advances in Neural Information Processing Systems (NeurIPS)},
  year      = {2023}
}

@inproceedings{feng2024binding,
  title     = {How do Language Models Bind Entities in Context?},
  author    = {Feng, Jiahai and Steinhardt, Jacob},
  booktitle = {International Conference on Learning Representations (ICLR)},
  year      = {2024}
}

@inproceedings{finlayson2021agreement,
  title     = {Causal Analysis of Syntactic Agreement Mechanisms in Neural Language Models},
  author    = {Finlayson, Matthew and Mueller, Aaron and Gehrmann, Sebastian and Shieber, Stuart and Linzen, Tal and Belinkov, Yonatan},
  booktitle = {Proceedings of the Annual Meeting of the Association for Computational Linguistics (ACL)},
  year      = {2021}
}

@article{linzen2016agreement,
  title   = {Assessing the Ability of {LSTMs} to Learn Syntax-Sensitive Dependencies},
  author  = {Linzen, Tal and Dupoux, Emmanuel and Goldberg, Yoav},
  journal = {Transactions of the Association for Computational Linguistics},
  volume  = {4},
  pages   = {521--535},
  year    = {2016}
}

@inproceedings{hooker2019roar,
  title     = {A Benchmark for Interpretability Methods in Deep Neural Networks},
  author    = {Hooker, Sara and Erhan, Dumitru and Kindermans, Pieter-Jan and Kim, Been},
  booktitle = {Advances in Neural Information Processing Systems (NeurIPS)},
  year      = {2019}
}

@inproceedings{hase2021ood,
  title     = {The Out-of-Distribution Problem in Explainability and Search Methods for Feature Importance Explanations},
  author    = {Hase, Peter and Xie, Harry and Bansal, Mohit},
  booktitle = {Advances in Neural Information Processing Systems (NeurIPS)},
  year      = {2021}
}

@inproceedings{lindner2023tracr,
  title     = {Tracr: Compiled Transformers as a Laboratory for Interpretability},
  author    = {Lindner, David and Kram{\'a}r, J{\'a}nos and Farquhar, Sebastian and Rahtz, Matthew and McGrath, Thomas and Mikulik, Vladimir},
  booktitle = {Advances in Neural Information Processing Systems (NeurIPS)},
  year      = {2023}
}

@inproceedings{gupta2024interpbench,
  title     = {{InterpBench}: Semi-Synthetic Transformers for Evaluating Mechanistic Interpretability Techniques},
  author    = {Gupta, Rohan and Arcuschin, Iv{\'a}n and Kwa, Thomas and Garriga-Alonso, Adri{\`a}},
  booktitle = {Advances in Neural Information Processing Systems (NeurIPS), Datasets and Benchmarks Track},
  year      = {2024}
}

@inproceedings{lee2018mahalanobis,
  title     = {A Simple Unified Framework for Detecting Out-of-Distribution Samples and Adversarial Attacks},
  author    = {Lee, Kimin and Lee, Kibok and Lee, Honglak and Shin, Jinwoo},
  booktitle = {Advances in Neural Information Processing Systems (NeurIPS)},
  year      = {2018}
}

@inproceedings{sun2022knn,
  title     = {Out-of-Distribution Detection with Deep Nearest Neighbors},
  author    = {Sun, Yiyou and Ming, Yifei and Zhu, Xiaojin and Li, Yixuan},
  booktitle = {International Conference on Machine Learning (ICML)},
  year      = {2022}
}

@article{radford2019gpt2,
  title   = {Language Models are Unsupervised Multitask Learners},
  author  = {Radford, Alec and Wu, Jeffrey and Child, Rewon and Luan, David and Amodei, Dario and Sutskever, Ilya},
  journal = {OpenAI technical report},
  year    = {2019}
}

@inproceedings{biderman2023pythia,
  title     = {Pythia: A Suite for Analyzing Large Language Models Across Training and Scaling},
  author    = {Biderman, Stella and Schoelkopf, Hailey and Anthony, Quentin and Bradley, Herbie and O'Brien, Kyle and Hallahan, Eric and Khan, Mohammad Aflah and Purohit, Shivanshu and Prashanth, USVSN Sai and Raff, Edward and Skowron, Aviya and Sutawika, Lintang and van der Wal, Oskar},
  booktitle = {International Conference on Machine Learning (ICML)},
  year      = {2023}
}

@misc{nanda2022transformerlens,
  title        = {{TransformerLens}},
  author       = {Nanda, Neel and Bloom, Joseph},
  year         = {2022},
  howpublished = {\url{https://github.com/TransformerLensOrg/TransformerLens}}
}

@article{wu2024reply,
  title   = {A Reply to {Makelov} et al. (2023)'s ``Interpretability Illusion'' Arguments},
  author  = {Wu, Zhengxuan and Geiger, Atticus and Huang, Jing and Arora, Aryaman and Icard, Thomas and Potts, Christopher and Goodman, Noah D.},
  journal = {arXiv preprint arXiv:2401.12631},
  year    = {2024}
}

@article{mcgrath2023hydra,
  title   = {The Hydra Effect: Emergent Self-repair in Language Model Computations},
  author  = {McGrath, Thomas and Rahtz, Matthew and Kram{\'a}r, J{\'a}nos and Mikulik, Vladimir and Legg, Shane},
  journal = {arXiv preprint arXiv:2307.15771},
  year    = {2023}
}

@misc{chan2022scrubbing,
  title        = {Causal Scrubbing: a Method for Rigorously Testing Interpretability Hypotheses},
  author       = {Chan, Lawrence and Garriga-Alonso, Adri{\`a} and Goldowsky-Dill, Nicholas and Greenblatt, Ryan and Nitishinskaya, Jenny and Radhakrishnan, Ansh and Shlegeris, Buck and Thomas, Nate},
  year         = {2022},
  howpublished = {AI Alignment Forum}
}

@inproceedings{li2024optimal,
  title     = {Optimal Ablations for Interpretability},
  author    = {Li, Maximilian and Janson, Lucas},
  booktitle = {Advances in Neural Information Processing Systems (NeurIPS)},
  year      = {2024}
}

@inproceedings{mueller2025mib,
  title     = {{MIB}: A Mechanistic Interpretability Benchmark},
  author    = {Mueller, Aaron and others},
  booktitle = {International Conference on Machine Learning (ICML)},
  year      = {2025}
}

@article{qwen2025qwen25,
  title={Qwen2.5 Technical Report},
  author={{Qwen Team}},
  journal={arXiv preprint arXiv:2412.15115},
  year={2024}
}

\appendix
\section{Implementation details}
\label{app:impl}

\paragraph{GPT-2 configurations.} Main runs: GPT-2 small, seed 0; sites at the subject token (layers 2 and 4) and the final token (layer 8); pathways at $T\in\{L{+}2,L{+}4\}$ with $T<12$; 24 planted units per configuration; five pathway settings (none, silent, hijack, distributed, in-range) for each of the two tasks; up to 6{,}000 prompt pairs per task; float32; all downstream layers. IOI (unmodified model): the S2 token (layers 4, 6 and 8) and the final token (layers 8 and 10), seeds 0 and 1, restricted DAS with the top 32 components excluded (seed 0) or the top 4 excluded (seeds 0 and 1). Mitigation runs: layer 2, $T=4$, 900 interventions each, $\lambda=5$, all four pathway families; seeds 0 and 1 for gender and seed 0 for SVA. These \NCfgVone{} configurations ran on one NVIDIA RTX 5060 Ti (16\,GB) and one Apple M3 Pro laptop; a configuration takes 2--8 minutes.

\paragraph{Larger models.} The three larger models use the four sites of \S\ref{sec:bench}, pathways at $T\in\{L{+}2,L{+}4\}$ (no-pathway configurations at $T{=}L{+}2$ only, which does not affect them), seeds 0--2, 24 planted units and bfloat16. Pythia-1.4B uses up to 6{,}000 prompt pairs and all downstream layers; Pythia-6.9B and Qwen2.5-7B use up to 4{,}000 prompt pairs and analyse every second downstream MLP layer, always including $T$. Each task--model pair has 108 configurations (4 sites $\times$ [4 families $\times$ 2 pathway layers $+$ 1 no-pathway] $\times$ 3 seeds). The unmodified-model sweep adds, for Pythia-6.9B and Qwen2.5-7B, restricted DAS with $k\in\{1,4,32\}$ on gender, SVA and IOI at the same sites and seeds (IOI answers are prompt-dependent; the final-token and subject sites refer to the final and S2 positions). The \NCfgHtw{} configurations (\NIntHtw{} interventions) ran on a single NVIDIA H20 (96\,GB) with PyTorch 2.11, CUDA 12.8 and TransformerLens 2.18; a configuration takes 2--12 minutes.

\paragraph{Shared hyperparameters.} 1{,}500 interventions per configuration; Algorithm~1 uses 32 natural evaluation prompts; DAS directions are trained for 150 steps (restricted: 300) with Adam at learning rate $10^{-2}$ and batch size 128, on one third of the pairs (at most 1{,}500). Interventions are evaluated on disjoint pairs. Planted-pathway gain: $\LD$ change of 3 per natural standard deviation of overshoot. The mitigation penalty reads downstream MLP activations after the planted drive has been added, so it sees exactly the activity the model computes with.

\paragraph{Natural regimes.} Quantiles are computed per class over all filtered benchmark prompts. Each configuration asserts that planting the pathways leaves every natural decision unchanged.

\paragraph{Additional runs.} \NCfgAsc{} further configurations (\NIntAsc{} interventions) ran on a Huawei Ascend 910B2 with the same detector code: GPT-2 seeds 1--2 at the subject token ($L{=}4$, $T{=}6$) and final token ($L{=}8$, $T{=}10$) for all gender settings and part of SVA, with 3{,}000 interventions, 300 DAS steps (restricted: 600), 8{,}000 prompts and 64 evaluation prompts for Algorithm~1; and an exploratory comparison of Qwen2.5-1.5B and Qwen2.5-1.5B-Instruct on gender (subject token, $L{=}6$, $T{=}8$, seeds 0--2) with 300 interventions, 60 DAS steps, 1{,}200 prompts and every fourth downstream layer. Because budgets differ, these are reported separately (Tables~\ref{tab:ascend_gpt2} and~\ref{tab:qwen_preview}) and are not pooled with the main results.

\paragraph{Reproducibility.} Every table and figure is generated from the saved per-intervention arrays by a single script. Each saved configuration records its parameters, model revision, software versions and file hashes.

\section{Changes in version 2}
\label{app:changes}
Version 1 reported GPT-2 small only, with one seed. Version 2 adds Pythia-1.4B, Pythia-6.9B and Qwen2.5-7B with three seeds each; the HPC-PCA variant; a sweep over the number of components excluded from restricted DAS in unmodified models, including IOI for Pythia-6.9B and Qwen2.5-7B; a positive-control table per model; a seed table, a cross-hardware reproduction check, GPT-2 seeds 1--2 and an exploratory instruction-tuning comparison. The GPT-2 results of version 1 are unchanged. The main qualitative conclusions of version 1 hold at the larger scales, with two updates: projecting onto the local natural manifold partly detects in-range pathways in larger models, and unmodified Qwen2.5-7B yields candidate wrong-mechanism successes under restricted DAS.

\section{Additional tables}

\begin{table*}[!tp]
\centering
\footnotesize
\setlength{\tabcolsep}{2.2pt}
\caption{\textbf{Detecting pathway-dominated interventions} (planted pathway supplies $>$50\% of the decision margin), all interventions. Mean AUROC over configurations (sites, pathway layers and, for the larger models, seeds 0--2) with $\geq$5 positives; best per row in bold. \#cfg: configurations used (of 5 per GPT-2 row and 24 per row for the other models). $^\dagger$Measures used by \citet{grant2026divergent}.}
\label{tab:auroc_all}
\begin{tabular}{lll rr cccc c cccc}
\toprule
 & & & & & \multicolumn{4}{c}{\emph{site-level}} & \emph{downstream} & \multicolumn{4}{c}{\emph{downstream + causal (ours)}} \\
\cmidrule(lr){6-9}\cmidrule(lr){10-10}\cmidrule(lr){11-14}
Model & Task & Pathway & \#cfg & \#pos & kNN & Maha. & LPCA$^\dagger$ & Alg.~1$^\dagger$ & Down-OM & HPC-S & HPC-L & HPC-PCA & HPC-Loc \\
\midrule
GPT-2 & Gender & silent & 5 & 3203 & 0.74 & 0.80 & 0.75 & 0.86 & 0.92 & \textbf{1.00} & \textbf{1.00} & \textbf{0.99} & 0.99 \\
GPT-2 & Gender & hijack & 5 & 3620 & 0.75 & 0.78 & 0.75 & 0.83 & 0.92 & 0.68 & \textbf{1.00} & \textbf{0.99} & \textbf{0.99} \\
GPT-2 & Gender & distrib. & 5 & 3144 & 0.75 & 0.80 & 0.75 & 0.86 & 0.91 & 0.69 & \textbf{1.00} & \textbf{0.99} & \textbf{0.99} \\
GPT-2 & Gender & in-range & 5 & 223 & 0.66 & \textbf{0.78} & 0.67 & 0.66 & 0.74 & 0.57 & 0.59 & 0.74 & 0.74 \\
GPT-2 & SVA & silent & 5 & 410 & 0.84 & 0.91 & 0.84 & 0.94 & 0.99 & \textbf{1.00} & \textbf{1.00} & \textbf{1.00} & 0.99 \\
GPT-2 & SVA & hijack & 5 & 241 & 0.80 & 0.90 & 0.81 & 0.94 & 0.97 & 0.61 & \textbf{1.00} & 0.98 & 0.97 \\
GPT-2 & SVA & distrib. & 5 & 444 & 0.80 & 0.91 & 0.80 & 0.95 & 0.98 & 0.75 & \textbf{1.00} & \textbf{1.00} & 0.99 \\
GPT-2 & SVA & in-range & 4 & 183 & 0.56 & 0.73 & 0.57 & 0.68 & \textbf{0.92} & 0.63 & 0.62 & 0.76 & 0.73 \\
\midrule
Pythia-1.4B & Gender & silent & 24 & 15625 & 0.68 & 0.70 & 0.68 & 0.83 & 0.90 & \textbf{1.00} & \textbf{1.00} & \textbf{1.00} & \textbf{0.99} \\
Pythia-1.4B & Gender & hijack & 24 & 15883 & 0.66 & 0.68 & 0.66 & 0.80 & 0.92 & 0.72 & \textbf{0.99} & \textbf{0.99} & \textbf{0.99} \\
Pythia-1.4B & Gender & distrib. & 24 & 12648 & 0.71 & 0.73 & 0.71 & 0.87 & 0.94 & 0.75 & \textbf{0.99} & \textbf{0.99} & \textbf{0.99} \\
Pythia-1.4B & Gender & in-range & 14 & 2326 & 0.55 & 0.49 & 0.55 & 0.55 & 0.65 & 0.51 & 0.80 & 0.89 & \textbf{0.90} \\
Pythia-1.4B & SVA & silent & 14 & 2264 & 0.78 & 0.60 & 0.79 & 0.92 & 0.92 & \textbf{1.00} & 0.99 & 0.98 & 0.98 \\
Pythia-1.4B & SVA & hijack & 14 & 2777 & 0.77 & 0.56 & 0.77 & 0.90 & 0.95 & 0.64 & \textbf{0.98} & 0.98 & 0.98 \\
Pythia-1.4B & SVA & distrib. & 12 & 2385 & 0.76 & 0.51 & 0.76 & 0.89 & 0.92 & 0.62 & \textbf{0.98} & 0.97 & 0.97 \\
Pythia-1.4B & SVA & in-range & 12 & 1281 & 0.76 & 0.47 & 0.76 & 0.78 & 0.75 & 0.51 & 0.67 & 0.89 & \textbf{0.91} \\
\midrule
Pythia-6.9B & Gender & silent & 24 & 20446 & 0.67 & 0.65 & 0.66 & 0.79 & 0.87 & \textbf{0.99} & \textbf{0.99} & \textbf{0.99} & \textbf{1.00} \\
Pythia-6.9B & Gender & hijack & 24 & 20185 & 0.65 & 0.63 & 0.65 & 0.74 & 0.87 & 0.69 & \textbf{0.99} & \textbf{0.99} & \textbf{0.99} \\
Pythia-6.9B & Gender & distrib. & 24 & 14755 & 0.68 & 0.66 & 0.68 & 0.80 & 0.89 & 0.75 & 0.98 & \textbf{0.98} & \textbf{0.99} \\
Pythia-6.9B & Gender & in-range & 24 & 1392 & 0.53 & 0.49 & 0.53 & 0.58 & 0.61 & 0.63 & 0.68 & 0.80 & \textbf{0.87} \\
Pythia-6.9B & SVA & silent & 21 & 7796 & 0.75 & 0.57 & 0.75 & 0.81 & 0.86 & \textbf{0.99} & \textbf{0.99} & 0.98 & 0.98 \\
Pythia-6.9B & SVA & hijack & 21 & 8136 & 0.77 & 0.58 & 0.77 & 0.80 & 0.92 & 0.60 & \textbf{0.98} & 0.97 & 0.98 \\
Pythia-6.9B & SVA & distrib. & 21 & 6557 & 0.75 & 0.54 & 0.75 & 0.83 & 0.88 & 0.61 & \textbf{0.95} & 0.94 & 0.94 \\
Pythia-6.9B & SVA & in-range & 20 & 2027 & 0.67 & 0.50 & 0.66 & 0.63 & 0.67 & 0.52 & 0.69 & 0.84 & \textbf{0.85} \\
\midrule
Qwen2.5-7B & Gender & silent & 21 & 6670 & 0.76 & 0.84 & 0.78 & 0.94 & 0.96 & \textbf{1.00} & 0.99 & 0.99 & 0.98 \\
Qwen2.5-7B & Gender & hijack & 21 & 6187 & 0.76 & 0.84 & 0.78 & 0.91 & 0.97 & 0.69 & \textbf{0.99} & 0.98 & 0.98 \\
Qwen2.5-7B & Gender & distrib. & 21 & 5367 & 0.80 & 0.84 & 0.80 & 0.95 & \textbf{0.97} & 0.63 & \textbf{0.97} & \textbf{0.97} & 0.96 \\
Qwen2.5-7B & Gender & in-range & 7 & 1512 & 0.70 & 0.67 & 0.71 & 0.66 & \textbf{0.75} & 0.50 & 0.64 & 0.73 & 0.67 \\
Qwen2.5-7B & SVA & silent & 20 & 3053 & 0.76 & 0.82 & 0.77 & 0.95 & 0.94 & \textbf{1.00} & 0.99 & 0.97 & 0.96 \\
Qwen2.5-7B & SVA & hijack & 22 & 3107 & 0.74 & 0.81 & 0.75 & 0.93 & 0.96 & 0.54 & \textbf{0.99} & 0.95 & 0.93 \\
Qwen2.5-7B & SVA & distrib. & 15 & 2369 & 0.72 & 0.77 & 0.72 & 0.92 & 0.94 & 0.51 & \textbf{0.96} & 0.93 & 0.92 \\
Qwen2.5-7B & SVA & in-range & 17 & 1903 & 0.73 & 0.78 & 0.73 & \textbf{0.84} & 0.82 & 0.50 & 0.66 & 0.76 & 0.78 \\
\bottomrule
\end{tabular}
\end{table*}

\begin{table*}[!tp]
\centering
\scriptsize
\setlength{\tabcolsep}{1.8pt}
\caption{Wrong-mechanism successes: pooled AUROC (scores rank-normalised within each configuration) with 95\% bootstrap confidence intervals over interventions (1{,}000 resamples).}
\label{tab:bootstrap}
\begin{tabular}{lll r cccccc}
\toprule
Model & Task & Pathway & \#pos & site LPCA & site Maha. & Alg.~1 & Down-OM & HPC-L & HPC-Loc \\
\midrule
GPT-2 & Gender & silent & 1099 & 0.44 [0.42, 0.46] & 0.64 [0.63, 0.66] & 0.61 [0.59, 0.63] & 0.65 [0.63, 0.66] & 0.70 [0.69, 0.72] & 0.70 [0.68, 0.71] \\
GPT-2 & Gender & hijack & 1393 & 0.41 [0.39, 0.42] & 0.60 [0.58, 0.62] & 0.55 [0.54, 0.57] & 0.60 [0.59, 0.62] & 0.77 [0.75, 0.79] & 0.75 [0.73, 0.77] \\
GPT-2 & Gender & distrib. & 832 & 0.43 [0.41, 0.45] & 0.63 [0.61, 0.65] & 0.57 [0.55, 0.59] & 0.59 [0.57, 0.61] & 0.68 [0.66, 0.70] & 0.68 [0.65, 0.70] \\
GPT-2 & Gender & in-range & 71 & 0.28 [0.21, 0.36] & 0.37 [0.30, 0.43] & 0.23 [0.18, 0.27] & 0.38 [0.31, 0.44] & 0.48 [0.39, 0.57] & 0.56 [0.50, 0.61] \\
GPT-2 & SVA & silent & 24 & 0.41 [0.33, 0.48] & 0.78 [0.73, 0.83] & 0.86 [0.78, 0.93] & 0.92 [0.89, 0.95] & 0.99 [0.99, 1.00] & 0.97 [0.93, 0.99] \\
GPT-2 & SVA & hijack & 14 & 0.46 [0.37, 0.55] & 0.81 [0.77, 0.85] & 0.85 [0.78, 0.92] & 0.94 [0.92, 0.96] & 0.99 [0.99, 1.00] & 0.96 [0.91, 0.99] \\
GPT-2 & SVA & distrib. & 27 & 0.39 [0.33, 0.45] & 0.77 [0.73, 0.81] & 0.69 [0.63, 0.75] & 0.93 [0.89, 0.95] & 0.96 [0.94, 0.99] & 0.90 [0.81, 0.97] \\
GPT-2 & SVA & in-range & 134 & 0.33 [0.30, 0.37] & 0.59 [0.55, 0.64] & 0.51 [0.48, 0.54] & 0.90 [0.89, 0.92] & 0.92 [0.89, 0.95] & 0.90 [0.87, 0.93] \\
\midrule
Pythia-1.4B & Gender & silent & 5543 & 0.35 [0.35, 0.36] & 0.63 [0.62, 0.64] & 0.64 [0.63, 0.65] & 0.58 [0.57, 0.59] & 0.79 [0.78, 0.79] & 0.80 [0.79, 0.80] \\
Pythia-1.4B & Gender & hijack & 5569 & 0.34 [0.34, 0.35] & 0.63 [0.62, 0.63] & 0.58 [0.57, 0.59] & 0.59 [0.58, 0.60] & 0.79 [0.79, 0.80] & 0.81 [0.81, 0.82] \\
Pythia-1.4B & Gender & distrib. & 3554 & 0.35 [0.34, 0.36] & 0.59 [0.59, 0.60] & 0.59 [0.58, 0.60] & 0.58 [0.57, 0.59] & 0.74 [0.74, 0.75] & 0.77 [0.76, 0.77] \\
Pythia-1.4B & Gender & in-range & 761 & 0.21 [0.20, 0.23] & 0.43 [0.41, 0.45] & 0.22 [0.20, 0.23] & 0.29 [0.27, 0.30] & 0.72 [0.70, 0.74] & 0.80 [0.78, 0.81] \\
Pythia-1.4B & SVA & silent & 912 & 0.29 [0.28, 0.31] & 0.47 [0.45, 0.48] & 0.77 [0.76, 0.78] & 0.64 [0.63, 0.65] & 0.83 [0.82, 0.84] & 0.83 [0.82, 0.84] \\
Pythia-1.4B & SVA & hijack & 1514 & 0.29 [0.28, 0.30] & 0.46 [0.45, 0.48] & 0.62 [0.61, 0.63] & 0.74 [0.73, 0.75] & 0.79 [0.78, 0.80] & 0.79 [0.78, 0.79] \\
Pythia-1.4B & SVA & distrib. & 1255 & 0.28 [0.27, 0.29] & 0.45 [0.44, 0.47] & 0.63 [0.62, 0.64] & 0.75 [0.74, 0.76] & 0.79 [0.78, 0.80] & 0.78 [0.77, 0.79] \\
Pythia-1.4B & SVA & in-range & 452 & 0.34 [0.31, 0.38] & 0.44 [0.42, 0.47] & 0.41 [0.39, 0.43] & 0.50 [0.48, 0.52] & 0.67 [0.65, 0.70] & 0.78 [0.76, 0.80] \\
\midrule
Pythia-6.9B & Gender & silent & 8491 & 0.40 [0.39, 0.41] & 0.57 [0.56, 0.58] & 0.64 [0.64, 0.65] & 0.57 [0.56, 0.58] & 0.73 [0.73, 0.74] & 0.74 [0.73, 0.74] \\
Pythia-6.9B & Gender & hijack & 8096 & 0.40 [0.39, 0.41] & 0.56 [0.55, 0.57] & 0.62 [0.61, 0.63] & 0.61 [0.60, 0.61] & 0.75 [0.75, 0.76] & 0.76 [0.75, 0.76] \\
Pythia-6.9B & Gender & distrib. & 6043 & 0.44 [0.43, 0.45] & 0.54 [0.53, 0.55] & 0.63 [0.62, 0.64] & 0.60 [0.59, 0.61] & 0.74 [0.73, 0.75] & 0.74 [0.73, 0.75] \\
Pythia-6.9B & Gender & in-range & 346 & 0.33 [0.31, 0.36] & 0.41 [0.39, 0.43] & 0.38 [0.36, 0.39] & 0.47 [0.45, 0.49] & 0.62 [0.58, 0.65] & 0.77 [0.74, 0.81] \\
Pythia-6.9B & SVA & silent & 3221 & 0.46 [0.45, 0.47] & 0.53 [0.52, 0.54] & 0.66 [0.65, 0.67] & 0.67 [0.67, 0.68] & 0.80 [0.79, 0.81] & 0.79 [0.79, 0.80] \\
Pythia-6.9B & SVA & hijack & 3520 & 0.45 [0.44, 0.46] & 0.53 [0.52, 0.54] & 0.64 [0.63, 0.65] & 0.73 [0.72, 0.73] & 0.82 [0.81, 0.82] & 0.81 [0.81, 0.82] \\
Pythia-6.9B & SVA & distrib. & 2958 & 0.45 [0.44, 0.46] & 0.51 [0.50, 0.52] & 0.68 [0.67, 0.69] & 0.73 [0.73, 0.74] & 0.82 [0.81, 0.83] & 0.81 [0.80, 0.82] \\
Pythia-6.9B & SVA & in-range & 836 & 0.49 [0.46, 0.51] & 0.49 [0.47, 0.51] & 0.41 [0.39, 0.42] & 0.52 [0.50, 0.54] & 0.64 [0.61, 0.66] & 0.73 [0.72, 0.75] \\
\midrule
Qwen2.5-7B & Gender & silent & 3634 & 0.26 [0.25, 0.28] & 0.58 [0.56, 0.60] & 0.73 [0.72, 0.75] & 0.68 [0.66, 0.70] & 0.88 [0.87, 0.89] & 0.87 [0.86, 0.88] \\
Qwen2.5-7B & Gender & hijack & 3363 & 0.28 [0.27, 0.30] & 0.57 [0.55, 0.59] & 0.55 [0.53, 0.57] & 0.72 [0.70, 0.73] & 0.82 [0.81, 0.84] & 0.80 [0.79, 0.82] \\
Qwen2.5-7B & Gender & distrib. & 2911 & 0.32 [0.30, 0.33] & 0.51 [0.50, 0.53] & 0.73 [0.71, 0.74] & 0.76 [0.74, 0.78] & 0.84 [0.83, 0.85] & 0.82 [0.81, 0.83] \\
Qwen2.5-7B & Gender & in-range & 509 & 0.31 [0.28, 0.33] & 0.49 [0.47, 0.52] & 0.32 [0.30, 0.35] & 0.45 [0.43, 0.48] & 0.41 [0.39, 0.44] & 0.50 [0.47, 0.52] \\
Qwen2.5-7B & SVA & silent & 1705 & 0.30 [0.28, 0.31] & 0.42 [0.41, 0.44] & 0.80 [0.79, 0.81] & 0.80 [0.79, 0.81] & 0.89 [0.88, 0.89] & 0.86 [0.85, 0.87] \\
Qwen2.5-7B & SVA & hijack & 1758 & 0.30 [0.29, 0.32] & 0.42 [0.40, 0.43] & 0.72 [0.71, 0.73] & 0.87 [0.86, 0.88] & 0.91 [0.90, 0.92] & 0.85 [0.84, 0.86] \\
Qwen2.5-7B & SVA & distrib. & 1173 & 0.27 [0.26, 0.28] & 0.42 [0.40, 0.43] & 0.78 [0.76, 0.79] & 0.86 [0.85, 0.87] & 0.89 [0.88, 0.90] & 0.88 [0.87, 0.89] \\
Qwen2.5-7B & SVA & in-range & 915 & 0.27 [0.25, 0.28] & 0.41 [0.39, 0.43] & 0.55 [0.53, 0.56] & 0.61 [0.59, 0.62] & 0.65 [0.63, 0.66] & 0.83 [0.82, 0.85] \\
\bottomrule
\end{tabular}
\end{table*}

\begin{table}[!htbp]
\centering
\scriptsize
\setlength{\tabcolsep}{2.5pt}
\caption{\textbf{Seed variation.} AUROC for wrong-mechanism successes, first averaged over the configurations of each seed, then summarised over seeds 0--2 as mean$\pm$s.d.\ (min--max).}
\label{tab:seeds}
\begin{tabular}{lll ccc}
\toprule
Model & Task & Pathway & site LPCA & HPC-L & HPC-Loc \\
\midrule
Pythia-1.4B & Gender & silent & 0.36$\pm$0.01 (0.34--0.38) & 0.86$\pm$0.01 (0.84--0.87) & 0.87$\pm$0.01 (0.85--0.88) \\
Pythia-1.4B & Gender & hijack & 0.35$\pm$0.01 (0.33--0.35) & 0.88$\pm$0.03 (0.84--0.90) & 0.90$\pm$0.02 (0.87--0.92) \\
Pythia-1.4B & Gender & distrib. & 0.38$\pm$0.04 (0.32--0.41) & 0.82$\pm$0.03 (0.78--0.86) & 0.85$\pm$0.03 (0.81--0.88) \\
Pythia-1.4B & Gender & in-range & 0.24$\pm$0.05 (0.19--0.31) & 0.70$\pm$0.02 (0.68--0.73) & 0.85$\pm$0.02 (0.82--0.88) \\
Pythia-1.4B & SVA & silent & 0.53$\pm$0.04 (0.47--0.58) & 0.92$\pm$0.03 (0.88--0.94) & 0.91$\pm$0.02 (0.88--0.93) \\
Pythia-1.4B & SVA & hijack & 0.50$\pm$0.08 (0.39--0.56) & 0.93$\pm$0.02 (0.90--0.95) & 0.92$\pm$0.02 (0.90--0.94) \\
Pythia-1.4B & SVA & distrib. & 0.50$\pm$0.05 (0.44--0.56) & 0.89$\pm$0.03 (0.86--0.92) & 0.87$\pm$0.06 (0.79--0.94) \\
Pythia-1.4B & SVA & in-range & 0.47$\pm$0.02 (0.46--0.50) & 0.65$\pm$0.05 (0.58--0.71) & 0.83$\pm$0.01 (0.83--0.85) \\
\midrule
Pythia-6.9B & Gender & silent & 0.35$\pm$0.02 (0.32--0.37) & 0.83$\pm$0.04 (0.77--0.88) & 0.83$\pm$0.04 (0.78--0.88) \\
Pythia-6.9B & Gender & hijack & 0.34$\pm$0.01 (0.33--0.35) & 0.87$\pm$0.01 (0.86--0.88) & 0.88$\pm$0.01 (0.87--0.90) \\
Pythia-6.9B & Gender & distrib. & 0.43$\pm$0.01 (0.42--0.45) & 0.82$\pm$0.05 (0.75--0.86) & 0.82$\pm$0.04 (0.76--0.85) \\
Pythia-6.9B & Gender & in-range & 0.37$\pm$0.04 (0.33--0.41) & 0.60$\pm$0.04 (0.56--0.65) & 0.74$\pm$0.02 (0.71--0.77) \\
Pythia-6.9B & SVA & silent & 0.60$\pm$0.04 (0.55--0.65) & 0.90$\pm$0.01 (0.88--0.91) & 0.89$\pm$0.01 (0.88--0.89) \\
Pythia-6.9B & SVA & hijack & 0.59$\pm$0.05 (0.52--0.62) & 0.91$\pm$0.02 (0.89--0.93) & 0.90$\pm$0.02 (0.88--0.92) \\
Pythia-6.9B & SVA & distrib. & 0.62$\pm$0.01 (0.61--0.64) & 0.92$\pm$0.02 (0.89--0.93) & 0.91$\pm$0.02 (0.88--0.92) \\
Pythia-6.9B & SVA & in-range & 0.54$\pm$0.02 (0.51--0.55) & 0.65$\pm$0.01 (0.64--0.67) & 0.75$\pm$0.03 (0.71--0.78) \\
\midrule
Qwen2.5-7B & Gender & silent & 0.21$\pm$0.03 (0.19--0.25) & 0.88$\pm$0.03 (0.84--0.92) & 0.87$\pm$0.04 (0.82--0.91) \\
Qwen2.5-7B & Gender & hijack & 0.26$\pm$0.09 (0.13--0.34) & 0.93$\pm$0.02 (0.90--0.95) & 0.91$\pm$0.06 (0.83--0.95) \\
Qwen2.5-7B & Gender & distrib. & 0.30$\pm$0.12 (0.20--0.46) & 0.94$\pm$0.01 (0.93--0.95) & 0.93$\pm$0.02 (0.91--0.95) \\
Qwen2.5-7B & Gender & in-range & 0.32$\pm$0.02 (0.29--0.35) & 0.39$\pm$0.07 (0.31--0.49) & 0.48$\pm$0.05 (0.44--0.55) \\
Qwen2.5-7B & SVA & silent & 0.21$\pm$0.05 (0.15--0.25) & 0.97$\pm$0.01 (0.95--0.98) & 0.93$\pm$0.01 (0.91--0.95) \\
Qwen2.5-7B & SVA & hijack & 0.26$\pm$0.07 (0.16--0.34) & 0.96$\pm$0.00 (0.96--0.96) & 0.90$\pm$0.01 (0.88--0.92) \\
Qwen2.5-7B & SVA & distrib. & 0.23$\pm$0.02 (0.19--0.25) & 0.95$\pm$0.04 (0.90--0.99) & 0.93$\pm$0.04 (0.87--0.96) \\
Qwen2.5-7B & SVA & in-range & 0.20$\pm$0.01 (0.19--0.21) & 0.64$\pm$0.04 (0.60--0.69) & 0.87$\pm$0.10 (0.73--0.94) \\
\bottomrule
\end{tabular}
\end{table}

\begin{table}[!htbp]
\centering
\footnotesize
\setlength{\tabcolsep}{3pt}
\caption{\textbf{Cross-hardware reproduction} of one GPT-2 configuration (gender, silent, subject token, $L{=}2$, $T{=}4$, seed 0) on the original consumer GPU and on the H20 used for the new models, with the same code and hyperparameters.}
\label{tab:p0}
\begin{tabular}{ll ccc}
\toprule
Label / subset & Detector & original & H20 & $\Delta$ \\
\midrule
pathway-dominated, all & site LPCA & 0.69 & 0.74 & +0.05 \\
pathway-dominated, all & site Maha. & 0.79 & 0.79 & -0.01 \\
pathway-dominated, all & Alg.~1 & 0.85 & 0.81 & -0.04 \\
pathway-dominated, all & Down-OM & 0.89 & 0.90 & +0.02 \\
pathway-dominated, all & HPC-L & 1.00 & 1.00 & -0.00 \\
pathway-dominated, all & HPC-Loc & 0.99 & 0.99 & -0.00 \\
\midrule
wrong-mechanism, successful & site LPCA & 0.32 & 0.26 & -0.05 \\
wrong-mechanism, successful & site Maha. & 0.73 & 0.63 & -0.10 \\
wrong-mechanism, successful & Alg.~1 & 0.72 & 0.51 & -0.21 \\
wrong-mechanism, successful & Down-OM & 0.74 & 0.70 & -0.04 \\
wrong-mechanism, successful & HPC-L & 0.86 & 0.76 & -0.10 \\
wrong-mechanism, successful & HPC-Loc & 0.86 & 0.75 & -0.11 \\
\midrule
wrong-mechanism successes & count & 246/1024 & 66/853 & \\
G success rate & \% & 98.6 & 99.5 & \\
H success rate & \% & 12.1 & 8.9 & \\
\bottomrule
\end{tabular}
\end{table}

\begin{table}[!htbp]
\centering
\scriptsize
\setlength{\tabcolsep}{2.2pt}
\caption{\textbf{GPT-2 seed replication.} AUROC for wrong-mechanism successes on the same sites (subject token $L{=}4,T{=}6$; final token $L{=}8,T{=}10$): seed 0 from the main GPT-2 runs (1{,}500 interventions, 150 DAS steps) and seeds 1--2 run separately on an Ascend 910B2 at a larger budget (3{,}000 interventions, 300 DAS steps, 8{,}000 prompts, 64 evaluation prompts for Algorithm~1). Because budgets differ, the two groups are shown side by side and not pooled. G path.: pathway-dominated share (\%) of DAS (G) successes. SVA seeds 1--2 are incomplete (5 of 20 configurations).}
\label{tab:ascend_gpt2}
\begin{tabular}{ll r cccc c r cccc c}
\toprule
 & & \multicolumn{6}{c}{seed 0 (main runs)} & \multicolumn{6}{c}{seeds 1--2 (Ascend)} \\ \cmidrule(lr){3-8}\cmidrule(lr){9-14}
Task & Pathway & \#cfg & LPCA & Alg.~1 & HPC-L & HPC-Loc & G path. & \#cfg & LPCA & Alg.~1 & HPC-L & HPC-Loc & G path. \\
\midrule
Gender & silent & 2 & 0.40 & 0.65 & 0.82 & 0.79 & 100.0 & 4 & 0.39 & 0.60 & 0.86 & 0.85 & 96.1 \\
Gender & hijack & 2 & 0.39 & 0.46 & 0.78 & 0.73 & 93.9 & 4 & 0.31 & 0.71 & 0.93 & 0.92 & 96.2 \\
Gender & distrib. & 2 & 0.41 & 0.75 & 0.88 & 0.88 & 98.9 & 4 & 0.35 & 0.76 & 0.91 & 0.91 & 92.1 \\
Gender & in-range & 2 & 0.41 & 0.29 & 0.50 & 0.53 & 36.1 & 4 & 0.39 & 0.30 & 0.58 & 0.60 & 40.9 \\
SVA & silent & 2 & -- & -- & -- & -- & 0.4 & 4 & -- & -- & -- & -- & 2.1 \\
SVA & hijack & 2 & -- & -- & -- & -- & 0.3 & 1 & -- & -- & -- & -- & 0.0 \\
\bottomrule
\end{tabular}
\end{table}

\begin{table}[!htbp]
\centering
\scriptsize
\setlength{\tabcolsep}{2.2pt}
\caption{\textbf{Exploratory: Qwen2.5-1.5B base vs.\ Instruct} (gender, raw completion prompts without chat template, subject token $L{=}6$, $T{=}8$, seeds 0--2; small budget: 300 interventions, 60 DAS steps, 1{,}200 prompts, downstream layers sampled every fourth). AUROC for wrong-mechanism successes, mean over configurations with $\geq$5 positives (-- if none). Last column: successful interventions with HPC-L $>1$. Instruct in-range has 2 and 3 positives, too few to score.}
\label{tab:qwen_preview}
\begin{tabular}{ll rrr ccccc r}
\toprule
Model & Pathway & \#cfg & \#succ. & \#wrong & kNN & Alg.~1 & Down-OM & HPC-L & HPC-Loc & HPC-L$>$1 \\
\midrule
Qwen2.5-1.5B & hijack & 3 & 447 & 116 & 0.37 & 0.64 & 0.66 & 0.85 & 0.85 & 283 \\
Qwen2.5-1.5B & none & 3 & 591 & -- & -- & -- & -- & -- & -- & 0 \\
Qwen2.5-1.5B-Instruct & hijack & 3 & 517 & 154 & 0.43 & 0.70 & 0.72 & 0.92 & 0.91 & 327 \\
Qwen2.5-1.5B-Instruct & in-range & 2 & 379 & 5 & -- & -- & -- & -- & -- & 0 \\
Qwen2.5-1.5B-Instruct & none & 3 & 599 & -- & -- & -- & -- & -- & -- & 1 \\
\bottomrule
\end{tabular}
\end{table}

\end{document}